\PassOptionsToPackage{table}{xcolor}
\documentclass[]{selfevolagent}

\usepackage{microtype}
\microtypesetup{expansion=false}
\usepackage{graphicx}
\usepackage{subcaption}
\usepackage{booktabs} 
\usepackage{hyperref}

\usepackage{amsmath}
\usepackage{amssymb}
\usepackage{mathtools}
\usepackage{amsthm}

\usepackage{latexsym}
\usepackage[T1]{fontenc}
\usepackage[utf8]{inputenc}
\usepackage{microtype}
\usepackage{amsmath}
\usepackage{booktabs}
\usepackage{multirow}
\usepackage{amssymb}
\usepackage{balance}
\usepackage{subcaption}

\usepackage{pifont}
\usepackage{makecell}
\usepackage{filecontents}
\usepackage{bbm}
\usepackage{pgfplots}
\usetikzlibrary{patterns}
\usepackage[inline,shortlabels]{enumitem}
\usepackage{natbib}
\usepackage{csquotes}
\usepackage{hyperref}
\usepackage{url}
\usepackage{svg}
\usepackage{bm}
\usepackage{graphics}
\usepackage{graphicx}
\usepackage{array}
\usepackage[english]{babel}
\usepackage{textcomp}
\usepackage{latexsym}
\usepackage{siunitx}
\usepackage[normalem]{ulem}
\useunder{\uline}{\ul}{}
\usepackage{array}
\usepackage{titletoc}
\usepackage{tikz}
\usepackage{arydshln}
\usepackage[most]{tcolorbox}
\usepackage{algorithm}
\usepackage{fontawesome5}
\usepackage{wrapfig}
\usetikzlibrary{tikzmark}
\makeatletter
\newcommand*\myfontsize{%
\@setfontsize\myfontsize{7}{8}%
}
\makeatother

\usepackage{romannum, setspace, algorithm, algpseudocode, enumitem, pifont, etoolbox, dsfont, fontawesome5, longtable, array, arydshln, bbding, multicol, multirow}

\newcommand{\websiteicon}{\raisebox{-1.5pt}{\includegraphics[height=1.03em]{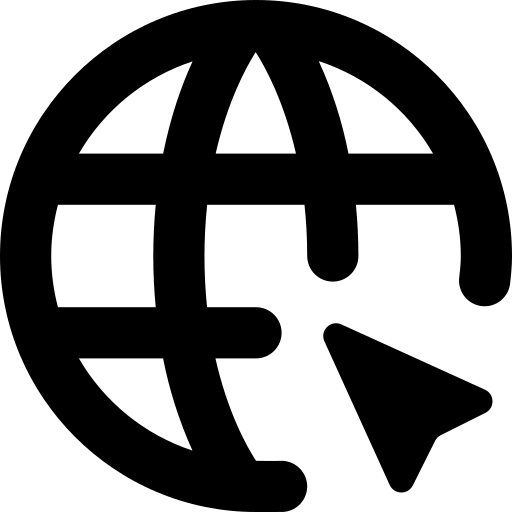}}}
\newcommand{\githubicon}{\raisebox{-1.5pt}{\includegraphics[height=1.03em]{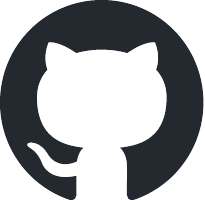}}}
\newcommand{\huggingfaceicon}{\raisebox{-1.5pt}{\includegraphics[height=0.96em]{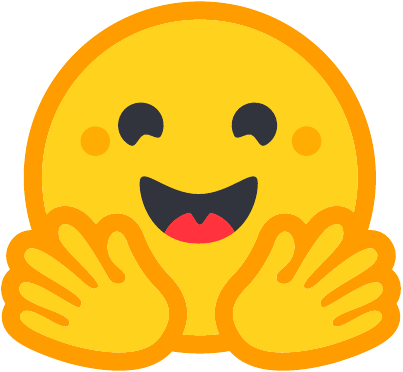}}}
\newcommand{\uiucicon}{\raisebox{-1.5pt}{\includegraphics[height=0.96em]{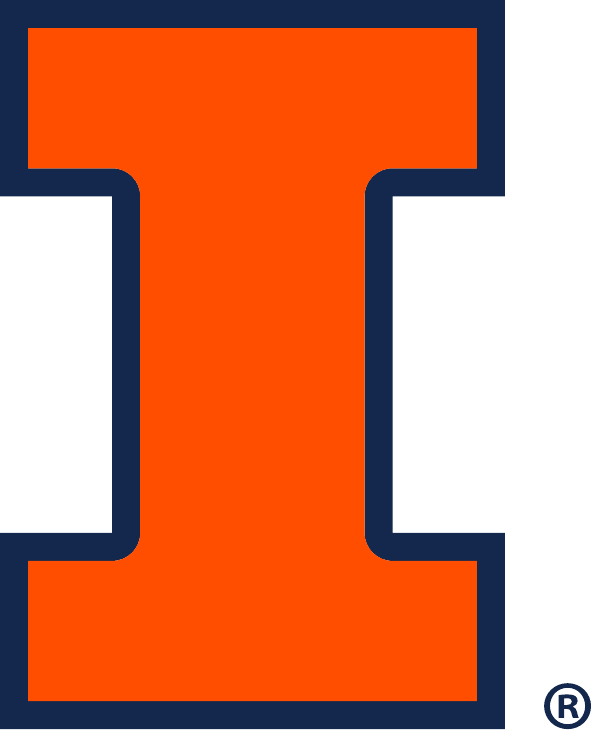}}}
\newcommand{\kaisticon}{\raisebox{-1.5pt}{\includegraphics[height=0.96em]{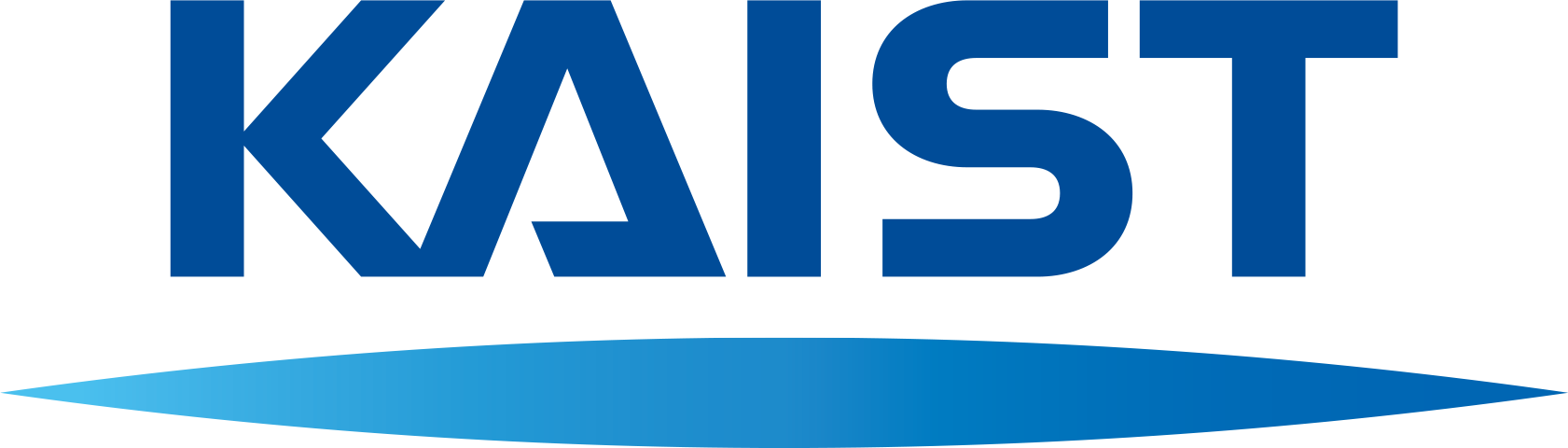}}}
\newcommand{\deepmindicon}{\raisebox{-1.5pt}{\includegraphics[height=0.96em]{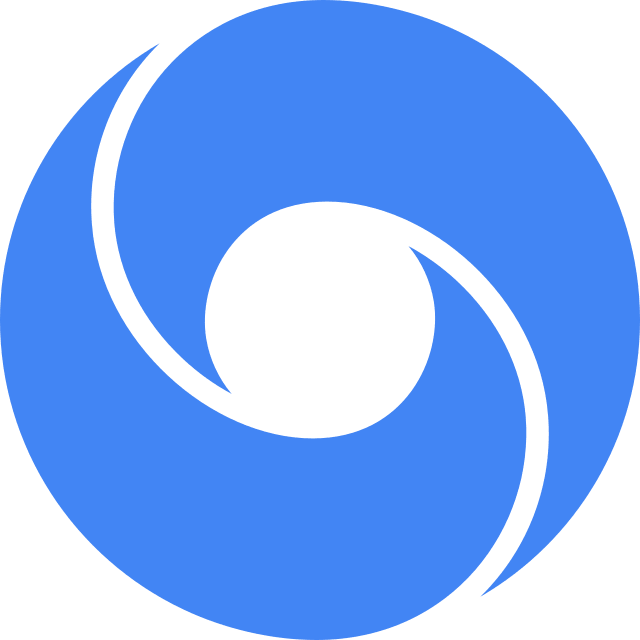}}}

\newcommand{\ie}{i.e.}
\newcommand{\eg}{e.g.}

\usepackage{amsmath,amsfonts,bm}

\def\eqref#1{equation~\ref{#1}}

\def\1{\bm{1}}

\DeclareMathAlphabet{\mathsfit}{\encodingdefault}{\sfdefault}{m}{sl}
\SetMathAlphabet{\mathsfit}{bold}{\encodingdefault}{\sfdefault}{bx}{n}

\usepackage{tcolorbox,titletoc}
\newcommand{\authcount}[1]{}
\usepackage{xstring,catchfile}

\usepackage{xcolor}

\title{STRIDE: When to Speak Meets Sequence Denoising for Streaming Video Understanding}
\author[1*]{Junho Kim}
\author[2*]{Hosu Lee}
\author[1]{James M. Rehg}
\author[3\dagger\ddagger]{Minsu Kim}
\author[2\dagger]{Yong Man Ro}

\affiliation[1]{\uiucicon~UIUC}
\affiliation[2]{\kaisticon~KAIST}
\affiliation[3]{\deepmindicon~Google DeepMind}
\contribution[*]{Equal contribution}
\contribution[\dagger]{Corresponding author}
\contribution[\ddagger]{Work done as an advisory role only.}

\metadata[Contact]{arkimjh@illinois.edu, leehosu01@kaist.ac.kr}
\metadata[\websiteicon~Project Page]{\href{https://interlive-team.github.io/STRIDE/}{https://interlive-team.github.io/STRIDE}}
\metadata[\huggingfaceicon~Huggingface]{\href{https://huggingface.co/interlive}{https://huggingface.co/interlive}}
\metadata[\githubicon~Code]{\href{https://github.com/interlive-team/STRIDE}{https://github.com/interlive-team/STRIDE}}

\abstract{Recent progress in video large language models (Video-LLMs) has enabled strong offline reasoning over long and complex videos. However, real-world deployments increasingly require streaming perception and proactive interaction, where video frames arrive online and the system must decide not only what to respond, but also when to respond. In this work, we revisit proactive activation in streaming video as a structured sequence modeling problem, motivated by the observation that temporal transitions in streaming video naturally form span-structured activation patterns. To capture this span-level structure, we model activation signals jointly over a sliding temporal window and update them iteratively as new frames arrive. We propose \textit{STRIDE} (Structured Temporal Refinement with Iterative DEnoising), which employs a lightweight masked diffusion module at the activation interface to jointly predict and progressively refine activation signals across the window. Extensive experiments on diverse streaming benchmarks and downstream models demonstrate that \textit{STRIDE} shows more reliable and temporally coherent proactive responses, significantly improving \textit{when-to-speak} decision quality in online streaming scenarios.}

\begin{document}
\pagenumbering{arabic}
\maketitle

\section{Introduction}
Along with recent advances in large language models (LLMs)~\cite{brown2020language,touvron2023llama,chatgpt,reid2024gemini,yang2025qwen3}, large vision-language models (LVLMs)~\cite{li2023blip, liu2023visual, dai2023instructblip, liu2023improved, chen2023internvl} have also achieved impressive performance across a wide range of image understanding and reasoning tasks. Building upon these advances, various video specialized models (\textit{i.e.,} Video-LLMs)~\cite{lin2023video,zhang2023video,kim2024salova,zhang2024long,li2025llama} further extend them to the temporal sequences, demonstrating remarkable capabilities in reasoning over video contents. However, existing Video-LLMs mostly operate in an offline manner, processing pre-recorded videos with access to the entire temporal context before generating responses. This fundamentally limits their capabilities to real-world streaming deployments such as egocentric assistants~\cite{huang2024vinci}, autonomous driving~\cite{xie2025glad}, or embodied AI agents~\cite{wei2025streamvln}, where the model must continuously perceive an ongoing video stream and decide \textit{when} and \textit{what} to respond in real time. 

Recognizing this gap, recent works have delved into streaming video understanding (SVU), where models continuously ingest incoming frames and maintain a temporal understanding on-the-fly~\cite{wang2024videollm,zhang2025flash,yang2025streammem,ning2025livevlm,yao2025timechat,zhang2026querystream}. Despite these advances, the approach is still \textit{reactive}, lacking a capability to determine when a response should be triggered. Expanding beyond the streaming scope, several works have explored \textit{proactive} response generation by leveraging special tokens~\cite{chen2024videollm,chen2025livecc,xu2025streamingvlm} to implicitly learn response timing or an agent-driven interaction approach~\cite{xiong2025streaming,yang2025streamagent}. More recently, several standalone activation modules~\cite{qian2024streaming,qian2025dispider,wang2025streambridge} have been proposed, especially those that decouple the streaming pipeline into two stages: a lightweight front-end that predicts activation signals at each frame to identify triggering moments, followed by a downstream Video-LLM that, when activated, consumes the accumulated frame cache to generate responses.

Within this decomposed framework, a straightforward way to train the activation module is to treat it as a binary classification problem as in~\cite{qian2024streaming,qian2025dispider,wang2025streambridge}, where at each time step a model predicts whether to trigger a response under binary supervision. However, such approach reduces activation to point-wise \texttt{0}/\texttt{1} decisions, answering ``\textit{should I respond now?}'' at each time step, without explicitly modeling how activation states transition across a temporal span. This often results in flickering activations and poorly resolved transition boundaries, causing unstable triggering behavior and fragmented activation spans. In practice, a reliable activation module must not only predict isolated labels, but also model how activation states change over time, capturing consistent \texttt{0}→\texttt{1} onsets, sustained \texttt{1}→\texttt{1} persistence, and well-resolved \texttt{1}→\texttt{0} offsets, so as to form coherent contiguous activation spans. In this sense, streaming and proactive triggering is more analogous to a span-structured decision rather than a point-wise one. To account for this span-level structure, an activation module should jointly model the activation sequence within a temporal neighborhood, so that the downstream Video-LLM can be activated under well-scoped visual context (neither prematurely with insufficient evidence nor too late after the moment has passed).

Motivated by recent advances in masked diffusion models~\cite{nie2025large,you2025llada,li2025lavida} (MDMs), which enable joint prediction over partially masked discrete sequences, we revisit streaming and proactive activation as \textit{structured sequence modeling} over an activation window. Unlike point-wise decision-making, masked diffusion operates on an entire sequence and iteratively refines corrupted states within context, naturally aligning with the span-structure of streaming trigger. Building on this, we propose \textit{STRIDE} (\textit{S}tructured \textit{T}emporal \textit{R}efinement with \textit{I}terative \textit{DE}noising), a proactive streaming framework that models the when-to-speak decision as structured sequence prediction, explicitly capturing span-level structure and activation state transitions. Specifically, during training, we employ boundary-aware span masking strategies that corrupt contiguous regions of the activation sequence, encouraging the model to reason about onset and offset from broader temporal context rather than relying on isolated binary signals. At inference time, as new frames arrive, \textit{STRIDE} progressively updates the activation window by carrying forward confident states and remasking uncertain positions, enabling temporally coherent span under partial observability while remaining plug-and-play and compatible with off-the-shelf Video-LLMs.

Through extensive experiments and comprehensive analyses on streaming benchmarks and downstream models, we corroborate that \textit{STRIDE} produces more reliable and temporally coherent proactive responses in online settings, significantly improving the \textit{when-to-speak} decisions. 

Our contributions can be summarized as follows:
\begin{itemize}
    \item We revisit proactive streaming activation in Video-LLMs and reformulate the when-to-speak problem as structured sequence modeling over a temporal activation window, establishing span-level activation as the prediction unit.
    \item We propose \textit{STRIDE} (\textit{S}tructured \textit{T}emporal \textit{R}efinement with \textit{I}terative \textit{DE} noising), a lightweight masked diffusion-based activation model that jointly predicts activation sequences and captures span-level structure.
    \item We validate \textit{STRIDE} through extensive experiments on diverse streaming benchmarks and downstream backbones, demonstrating more stable proactive triggering and improved temporal consistency in online settings.
\end{itemize}
\section{Related Work}

\subsection{Large Vision-Language Models}
Early works on LVLMs~\cite{liu2023visual,dai2023instructblip,li2024llava} have demonstrated that visual instruction tuning, which pairs a vision encoder with a LLM backbone and trains on instruction-following data, can output strong general-purpose capabilities for back-and-forth multi-modal conversation. Subsequent efforts~\cite{chen2023internvl,wang2024enhancing,zhu2025internvl3,wang2025internvl3} have focused on scaling model and data, improving visual tokenization, and aligning vision and language representations at scale. Especially, Qwen families~\cite{bai2023qwen,wang2024qwen2,bai2025qwen2,Qwen3-VL} improve visual processing efficiency and capability with dynamic resolution and stronger multi-modal pretraining, enabling more robust perception and reasoning over complex visual inputs. In addition, Video-LLMs~\cite{zhang2023video,li2024mvbench,song2024moviechat,zhang2024long} extend its scope to temporal understanding by treating video as a sequence of images, introducing video-specific connector~\cite{lin2023video,kim2024salova,zhang2025videollama} and training pipelines~\cite{li2024llavanext, sharegemini, zhang2024video} that better capture spatiotemporal dynamics, thereby leading to stronger performance on video QA and captioning tasks. Despite these advances, most LVLMs remain confined to an \textit{offline} setting, where the entire video clip is available prior to inference, limiting their applicability in real-time streaming scenarios.

\subsection{Streaming Video Understanding}
A growing body of works~\cite{qian2024streaming,zhang2025eyes,li2025lion} has explored expanding video understanding into the \textit{streaming} regime, where frames arrive online and frameworks must maintain state over time. One line of research adapts models to streaming interaction by redesigning training objectives and data formats for continuous inputs~\cite{chen2024videollm}, incorporating memory-augmented architectures for multi-turn streaming~\cite{zhang2025flash,xiong2025streaming}, and leveraging real-time commentary pipelines that integrate video speech transcripts with instruction tuning~\cite{chen2025livecc,xu2025streamingvlm}. Another branch emphasizes efficiency for unbounded video streams through memory aggregation for long streams~\cite{zhang2025flash}, streaming-aligned KV-cache strategies~\cite{xu2025streamingvlm, ning2025livevlm, yang2025streammem}, and redundant visual token dropping based on inter-frame similarity~\cite{yao2025timechat}. While these approaches have enabled Video-LLMs to process continuous streams, they remain fundamentally \textit{reactive}, generating responses only upon instantaneous user queries.

Addressing this gap, another direction tackles the \textit{proactive response}, which targets deciding \textit{when to respond} as the video unfolds. Several approaches exploit \texttt{EOS} token within autoregressive generation to implicitly determine response timing~\cite{chen2024videollm, xu2025streamingvlm}, conflating the triggering with language generation. Agentic methods explicitly model task-relevant temporal intervals for goal-driven triggering~\cite{yang2025streamagent}, or query-aware visual pruning with proactive response mechanisms~\cite{zhang2026querystream}. Most relevant to our work, recent modular approaches~\cite{qian2024streaming, qian2025dispider, wang2025streambridge, wang2024videollm} explicitly decouple the pipeline into a lightweight front-end that predicts per-frame binary activation signals and a downstream Video-LLM that generates responses upon triggering. While such a modular design preserves the downstream Video-LLM’s capabilities, reducing activation to point-wise binary supervision undermines the temporal coherence of contiguous activation spans. In this work, we retain the modular design while recasting activation as a \textit{structured sequence prediction} problem, leveraging masked diffusion to jointly model activation sequences over a temporal window and capture span-level temporal coherence.

\subsection{Discrete Diffusion Language Models}
Recent progress in discrete diffusion language models (dLLMs)~\cite{nie2025large, sahoo2024simple, lou2023discrete} revisits diffusion as an alternative to autoregressive decoding for text generation using masked diffusion models mechanism. Instead of generating tokens strictly left-to-right,
dLLMs iteratively denoise masked token sequences, enabling bidirectional conditioning and parallel token updates, which naturally supports controllable generation. Subsequent efforts have further scaled dLLMs by converting pretrained autoregressive models into diffusion-based counterparts~\cite{gong2024scaling, ye2025dream}, and improved their alignment and inference efficiency through parallel decoding strategies~\cite{chen2025dparallel}. Especially, LLaDA series scale masked diffusion to large LLMs~\cite{nie2025large} and further explores post-training alignment~\cite{zhu2025llada} as well as system-level scaling by converting pretrained AR models into diffusion models~\cite{bie2025llada2},
thereby inheriting knowledge while retaining the non-autoregressive generation benefits. This research scope has also been extended to the multi-modal setting, where vision encoders are coupled with diffusion language backbones for visual instruction following~\cite{li2025lavida, you2025llada, yu2025dimple, cheng2025sdar}, demonstrating that dLLMs can benefit from parallel decoding and bidirectional reasoning in vision-language tasks. Different from these works that primarily replace the autoregressive decoder for textual response generation, our work leverages masked diffusion for proactive streaming activation. We treat the \textit{when-to-speak} signal as a structured discrete activation sequence over a temporal window, jointly predicting the activation states for the incoming video streams.
\section{Proposed Method}
\label{sec:method}

\subsubsection{Preliminaries: Masked Diffusion Models.}
Recently, diffusion language models (dLLMs)~\cite{nie2025large, zhu2025llada, li2025lavida, you2025llada} have shown remarkable progress as an alternative paradigm to autoregressive language modeling, replacing left-to-right token generation with a masked diffusion process that iteratively denoises discrete token sequences. Given a sequence of $L$ tokens $\mathbf{x}_0 = (x_0^1, \ldots, x_0^L)$, the forward process progressively corrupts $\mathbf{x}_0$ by independently replacing each token with a mask token $\texttt{[M]}$ with probability $t \in [0,1]$, generating a partially masked sequence $\mathbf{x}_t$. At $t=0$ the sequence is fully observed, while at $t=1$ it is entirely masked.

\begin{figure*}[t!]
\centering
\includegraphics[width=1.0\textwidth]{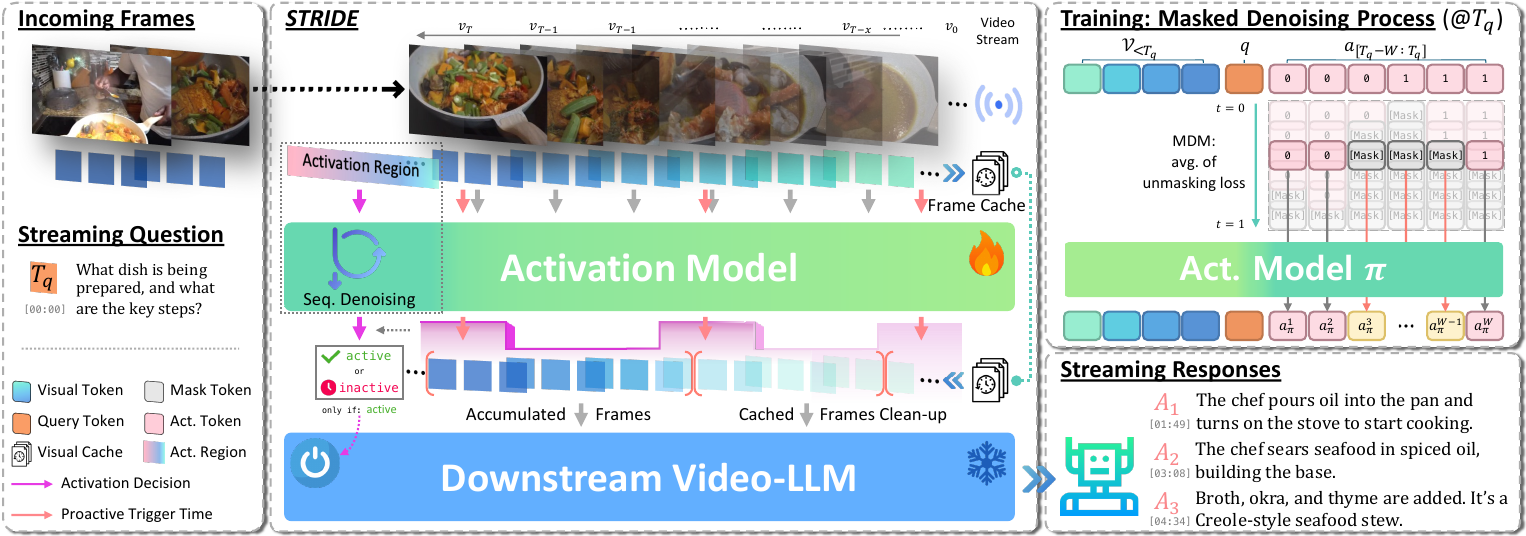}
\caption{Overview of \textbf{\textit{STRIDE}}, which operates in a streaming setting where frames arrive online. A lightweight activation model based on masked diffusion maintains an activation region over a sliding temporal window and iteratively denoises masked activation states to predict a coherent trigger segment. A trigger is issued only if an active span is sustained for a predefined span ratio. When activation is triggered, the accumulated frame context is forwarded to a downstream Video-LLM to generate the response.}
\label{fig:1}
\end{figure*}

The core of MDMs is a \textit{mask predictor} $p_\theta(\cdot \mid \mathbf{x}_t)$ with bidirectional attention that takes $\mathbf{x}_t$ as input and predicts all masked tokens simultaneously. The reverse process~\cite{austin2021structured, shi2024simplified, sahoo2024simple} recovers $\mathbf{x}_0$ from $\mathbf{x}_t$ by iteratively applying this mask predictor, which is trained by minimizing a cross-entropy loss computed only over the masked positions:
\begin{equation}
    \mathcal{L}(\theta) = -\mathbb{E}_{t, \mathbf{x}_0, \mathbf{x}_t} \left[ \frac{1}{t} \sum_{i=1}^{L} \mathds{1}[x_t^i = \texttt{M}] \log p_\theta(x_0^i \mid \mathbf{x}_t) \right],
    \label{eq:mdm_loss}
\end{equation}
where $t \sim U[0,1]$ and $\mathbf{x}_t$ is sampled from the forward process. This serves as an upper bound on the negative log-likelihood of the model distribution~\cite{nie2025large,bie2025llada2}.

At inference, generation proceeds by initializing a fully masked sequence $\mathbf{x}_1$ and simulating the reverse process through $K$ discrete steps decreasing from $t = 1 \to 0$. At each step, the mask predictor predicts all masked positions, and a subset of predictions is accepted while the remaining positions are remasked for subsequent refinement. This iterative predict-and-refine procedure enables MDMs to generate coherent sequences through progressive unmasking with bidirectional context. 

\subsection{STRIDE: Proactive Streaming Framework}
\subsubsection{Problem Formulation.}
The proposed \textit{STRIDE} (shown in \cref{fig:1}) considers the streaming video understanding setting where a model continuously processes video streams $\mathcal{V}{=}\{v_1, v_2, \ldots, v_T, \ldots\}$ with $v_T$ denoting the incoming visual frame arriving at time step $T$, interleaved with user queries and model-generated responses over time. Unlike offline Video-LLMs that have access to the holistic video sequences before generating a response, a streaming model must work under partial observability, where only the frames observed so far $\mathcal{V}_{\leq T}{=}\{v_1, \ldots, v_T\}$ and context priors $\mathcal{C}_T$ (\textit{e.g.,} user query $q$ and prior interaction history) are available. At every time step $T$, the model faces two sequential decisions: (\lowercase\expandafter{\romannumeral1}) \textit{whether} to respond, and (\lowercase\expandafter{\romannumeral2}) if so, \textit{what} to respond. \textit{STRIDE} adopts a two-stage streaming framework to decouple these decisions.

\subsubsection{Two-Stage Architecture.}
As illustrated in~\cref{fig:1}, \textit{STRIDE} is designed with the two-stage streaming framework. A lightweight \textit{Activation Model} $\pi$ continuously monitors the incoming stream and determines whether a proactive response should be triggered. Once a response is triggered at time step $T$, the accumulated visual context since the most recent query time $T_q$, denoted $\mathcal{V}_{[T_q:T]}$, together with the interaction context $\mathcal{C}_T$, is forwarded to a downstream Video-LLM, which generates the response $\mathcal{R}_T = f(\mathcal{C}_T, \mathcal{V}_{[T_q:T]})$. The generated response $\mathcal{R}_T$ is appended to the interaction context, updating it to $\mathcal{C}_{T'} = \mathcal{C}_T \cup \mathcal{R}_T$, enabling awareness of prior responses and maintaining dialogue coherence across multiple activation events. After each triggered response, the visual accumulation is cleared and restarted from the current time step, ensuring that subsequent activation decisions operate on fresh streaming context. This modular design cleanly separates \textit{when-to-speak} modeling from downstream response generation.

\subsubsection{Span-Level Activation Modeling.}
To formalize the activation decision, we represent activation as a window-level sequence of size $W$ anchored at time step $T$, and model it as a sequence-level prediction over this temporal window. Specifically, we define an activation region $\textbf{\textit{a}}_T = [a_{T-W}, \ldots, a_T] \in \{\texttt{0},\texttt{1}\}^W$, indicating inactive or active states within the window. This windowed formulation enables the activation model to learn contiguous activation spans and their transition dynamics (\texttt{0}→\texttt{1} onset, \texttt{1}→\texttt{1} persistence, \texttt{1}→\texttt{0} offset), aligning the prediction unit with span-level structures rather than isolated point-wise decisions.

As the video stream unfolds, incoming frames are sampled at 1 FPS and encoded into visual tokens by a vision encoder, which are accumulated in a running visual cache. At each time step $T$, the activation region $\textbf{\textit{a}}_T$ is appended after the visual cache as the prediction target. Each activation token takes values from the discrete vocabulary $\{\texttt{0}, \texttt{1}, \texttt{[M]}\}$, where $\texttt{[M]}$ denotes masked positions to be denoised. The activation model conditions on the visual cache and jointly infers masked activation states within the temporal window. When the activation state is determined to be active under the span-based criterion, the accumulated visual context is forwarded to the downstream Video-LLM for response generation.

\begin{figure*}[t]
\centering
\includegraphics[width=1.0\textwidth]{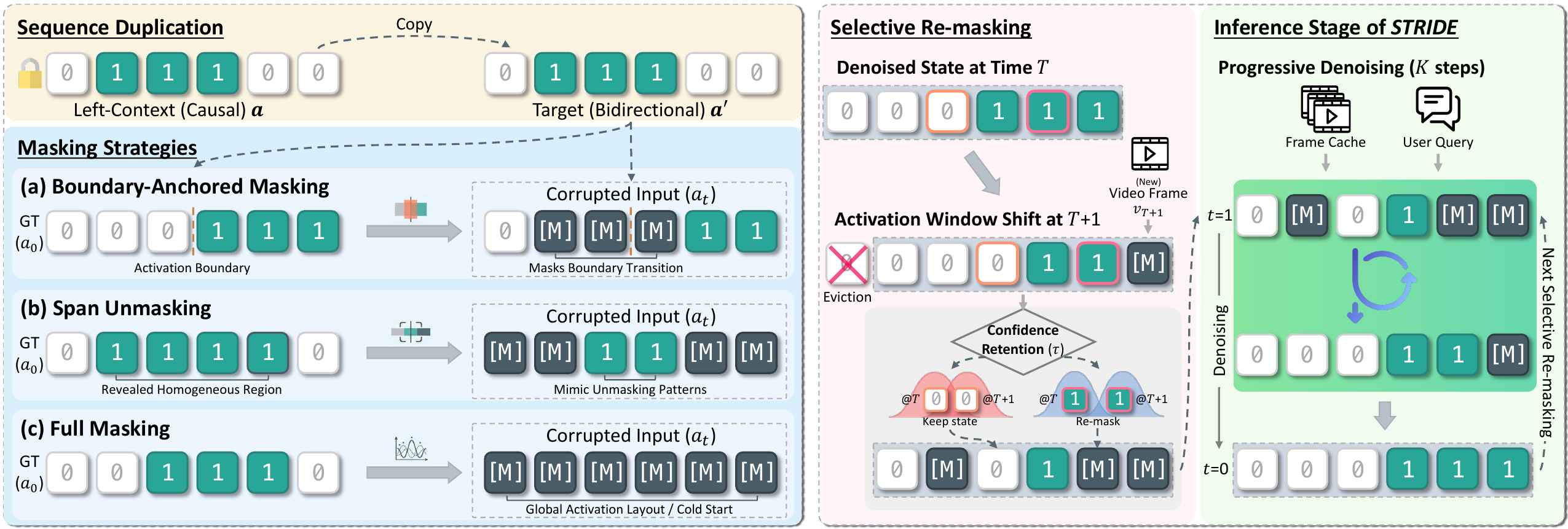}
\caption{Activation modeling and inference stage of \textbf{\textit{STRIDE}}. Training applies sequence duplication and three masking strategies (boundary-anchored masking, span unmasking, full masking). During inference, the activation window slides with incoming frames, retaining confident past decisions while selectively re-masking and progressively denoising uncertain positions.}
\label{fig:2}
\end{figure*}

\subsection{Training: Activation as Sequence Denoising}
\label{sec:training}
\subsubsection{Structured Masking Strategies for Activation Denoising.}
\label{sec:masking}
To train the activation model under the structured formulation, we propose a mixture of three corruption strategies instead of the standard MDM~\cite{nie2025large}, which samples mask positions independently. Such masking is inappropriate for our activation learning as the target sequence consists of contiguous active regions; isolated unmasked tokens between active positions make the denoising task trivially solvable through local interpolation, bypassing the need for genuine temporal understanding. The proposed masking mixture shown in \cref{fig:2} (left) is composed of: 
\begin{itemize}[left=0pt]
    \item \textbf{Boundary-Anchored Span Masking} masks a contiguous block overlapping with at least one activation boundary, forcing the model to determine where the active region begins and ends from broader temporal context.
    \item \textbf{Span Unmasking} starts from a fully masked sequence and reveals a contiguous block while keeping boundary-adjacent positions masked, mimicking the inference-time pattern where high-confidence tokens are unmasked consecutively in homogeneous regions. 
    \item \textbf{Full Masking} initially masks the entire activation sequence (cold-start) to stabilize the reverse step by training the model to estimate the global activation layout from visual context alone.
\end{itemize}

During training, each sample is randomly corrupted using one of three structured masking strategies, each selected with equal probability. These structured strategies encourage the model to reason over contiguous activation spans and their boundary transitions, rather than relying on isolated token predictions. As a result, the activation module learns span-level consistency that better aligns with the sequential and partial observability of streaming proactive triggering.

\subsubsection{Recovering Bidirectional Conditioning with Sequence Duplication.} Ma-sked diffusion predicts masked positions using full-sequence context, whereas an AR-pretrained activation model is trained with causal attention that only exposes left context. We therefore introduce an input reparameterization that enables bidirectional conditioning without altering the underlying causal attention layers. Specifically, we employ \textit{sequence duplication}, appending a copy of the activation region to form $[\textbf{\textit{a}},\; \textbf{\textit{a}}']$, where the copy carry identical activation tokens but serve distinct roles. The duplicated sequence $\textbf{\textit{a}}'$ produces diffusion predictions, while $\textbf{\textit{a}}$ serves as a conditioning prefix under causal attention. Concretely, since $\textbf{\textit{a}}$ is entirely placed before $\textbf{\textit{a}}'$, every token in $\textbf{\textit{a}}'$ can access all positions of $\textbf{\textit{a}}$ as left-context, providing full-window visibility for denoising without modifying the causal attention mask.

\subsubsection{Training Objective.}
Following the denoising process in~\cref{eq:mdm_loss}, we train the activation module by minimizing the masked cross-entropy loss over $\textbf{\textit{a}}'$, conditioned on the user query $q$ and the visual cache $\mathcal{V}_{\leq T}$:
\begin{equation}
    \mathcal{L}(\theta) = -\mathbb{E}_{t,\, \textbf{\textit{a}}'_0,\, \textbf{\textit{a}}'_t} \left[ \frac{1}{t} \sum_{j=1}^{W} \mathds{1}[a'^{j}_{t} = \texttt{M}] \log p_\theta(a'^{j}_{0} \mid q,\, \mathcal{V}_{\leq t},\, \textbf{\textit{a}}'_t) \right],
    \label{eq:stride_loss}
\end{equation}
where $\textbf{\textit{a}}'_0$ is the ground-truth activation sequence, $\textbf{\textit{a}}'_t$ is obtained by applying aforementioned our masking strategies at noise level $t \sim U[0,1]$, and the user query $q$ along with the visual cache $\mathcal{V}_{\leq t}$ serves as a fixed conditioning prefix, analogous to the prompt in the supervised fine-tuning of dLLMs~\cite{nie2025large, li2025lavida}.

\subsection{Inference: Streaming as Progressive Unmasking}
\label{stride_infder}
At inference time, \textit{STRIDE} maintains a sliding activation window and performs progressive refinement as illustrated in~\cref{fig:2} (right); confident past decisions are preserved, while uncertain and newly introduced positions are jointly refined with masked diffusion. Concretely, new time step $T\!+\!1$ proceeds in two stages:

\noindent(\lowercase\expandafter{\romannumeral1}) Selective Re-masking:
The activation sequence of size $W$ is shifted forward so that the region falling outside the window is evicted and a new frame is appended, causing the activation sequence to advance in time. The fully resolved activation $a_T^{j+1}$ previously assigned to position $j\!+\!1$ at time $T$ now maps to position $j$ at time $T\!+\!1$. To determine whether each carried-forward decision remains reasonable given the new visual evidence $v_{T+1}$, we apply a confidence-based retention: if $p_\theta(a_{T+1}^{j} \!=\! a_T^{j+1} \mid q,\, \mathcal{V}_{\leq T+1},\, \textbf{\textit{a}}_{T+1}) > \tau$, position $j$ inherits its previous decision; otherwise, it is re-masked to $\texttt{[M]}$ so that uncertain positions re-enter the denoising process alongside the newly appended slots.

\noindent(\lowercase\expandafter{\romannumeral2}) $K$-Step Progressive Denoising:
The masked positions obtained from the previous stage, comprising both newly appended slots and low-confidence re-masked slots, are resolved over $K$ denoising steps by prioritizing high-confidence positions first. At each step, the model computes the activation probability $p^{j} \!=\! p_\theta(a^j \!=\! 1 \mid q,\, \mathcal{V}_{\leq T+1},\, \textbf{\textit{a}}_{T+1})$ for every masked position and derives a confidence score $c^{j} \!=\! \max(p^{j},\, 1\!-\!p^{j})$, which measures how strongly the prediction leans toward either triggering or not. The top-$k$ positions ranked by $c^{j}$ are unmasked, where $k \!=\! \lceil N_{\text{init}} / K \rceil$ and $N_{\text{init}}$ is the number of masked positions established in stage (\lowercase\expandafter{\romannumeral1}), while the rest remain masked for subsequent refinement. By revealing high-confidence decisions first, this schedule establishes reliable temporal anchors that progressively stabilize the remaining ambiguous boundary regions.

After $K$ steps, the activation window is fully resolved. A trigger at time $T\!+\!1$ is issued only if an active span is sustained for at least $\gamma$ consecutive positions, where $\gamma$ denotes the required span ratio for triggering.
\section{Experiments}
\label{sec:experiments}
\subsection{Experimental Setting}

\subsubsection{Implementation \& Training Details.}
The activation model is initialized from a compact vision-language model using Qwen3-VL-2B~\cite{Qwen3-VL} to minimize streaming overhead. The downstream Video-LLMs are kept frozen, ensuring full modularity between the two stages. Incoming video frames are sampled at 1 FPS and encoded into the visual cache as the stream progresses. For the denoising process, we adopt the low-confidence remasking strategy~\cite{nie2025large} with $K{=}8$ sampling steps during inference. $\tau$ is set to 0.75, and $\gamma$ is set to 1 following the benchmark evaluation protocol. The entire activation model is trained on a single node of 8 NVIDIA H100 GPUs,  while evaluation is conducted on a single H100 GPU. Comprehensive hyperparameter settings and additional configurations are provided in the Appendix.

For the training data, we curate a diverse collection of temporally annotated video datasets spanning multiple video understanding tasks, including dense video captioning~\cite{caba2015activitynet,liu2024bench,huang2024lita}, temporal activity detection~\cite{sigurdsson2016hollywood,sigurdsson2018charades}, grounded video QA~\cite{wang2024grounded}, sequential step recognition~\cite{zhou2018towards}, and moment localization~\cite{anne2017didemo}. We convert each temporal annotation into a binary activation sequence aligned with the frame sampling rate, where frames within annotated spans are labeled as active (\texttt{1}) and the remaining frames as inactive (\texttt{0}).

\subsubsection{Benchmarks \& Baselines.}
We evaluate \textit{STRIDE} on three complementary benchmarks. OVO-Bench~\cite{niu2025ovo} assesses online video understanding across backward tracing, real-time visual perception, and forward active responding, where the model must delay its response until sufficient future evidence is available. StreamingBench~\cite{lin2024streamingbench} evaluates streaming comprehension through 18 tasks spanning real-time visual understanding, omni-source understanding, and contextual understanding including proactive output and sequential question answering. In addition, we evaluate on subsets of ET-Bench~\cite{liu2024bench}, including Temporal Video Grounding (TVG), Episodic Memory (EPM), Temporal Action Localization (TAL), Dense Video Captioning (DVC), and Step Localization and Captioning (SLC). This setup evaluates activation timing precision by measuring how accurately the model identifies event boundaries. For baselines, we compare against various offline Video-LLMs, online streaming proactive models~\cite{chen2024videollm,qian2025dispider,wang2025streambridge}, and proprietary models~\cite{reid2024gemini,gpt4o}. In addition, we include \textbf{Baseline-AR}, which serves as the primary counterpart to \textit{STRIDE}. Baseline-AR follows the same architecture and training setup as our method but replaces the masked diffusion activation module with an autoregressive binary prediction head trained with BCE loss, following the activation formulation described in prior work~\cite{wang2025streambridge}. This setup isolates the activation modeling strategy, enabling a direct comparison between masked denoising and autoregressive binary prediction.

\definecolor{headerblue}{HTML}{ECF4FF}
\definecolor{headergray}{HTML}{F2F2F2}
\definecolor{deltagreen}{HTML}{7CC472}
\newcommand{\gain}[1]{}

\begin{table*}[t]
\centering
\small
\setlength{\tabcolsep}{5pt}
\caption{Evaluation results on OVO-Bench~\cite{niu2025ovo}. Baseline-AR uses autoregressive binary prediction. Offline models follow the original single-turn protocol with segmented clips, whereas streaming methods process frames sequentially.}
\label{tab:ovo_bench_full}
\vspace{-2mm}
\resizebox{1.0\linewidth}{!}{
    \begin{tabular}{@{}l c cccccc c !{\vrule} ccc c !{\vrule} ccc c !{\vrule} c@{}}
\Xhline{2\arrayrulewidth}
\multirow{2}{*}[-4pt]{Method} & \multirow{2}{*}[-4pt]{\shortstack{\# of\\Frames}} & \multicolumn{7}{c}{\textbf{Real-Time Visual Perception}} & \multicolumn{4}{c}{\textbf{Backward Tracing}} & \multicolumn{4}{c}{\textbf{Forward Act. Responding}} & \textbf{Overall} \\
\cmidrule(l{2pt}r{2pt}){3-9} \cmidrule(l{2pt}r{2pt}){10-13} \cmidrule(l{2pt}r{2pt}){14-17} \cmidrule(l{2pt}r{2pt}){18-18}
& & OCR & ACR & ATR & STU & FPD & OJR & Avg. & EPM & ASI & HLD & Avg. & REC & SSR & CRR & Avg. & Avg. \\
\hline
\rowcolor{headergray} \multicolumn{18}{l}{\textit{Human}} \\
Human & - & 93.96 & 92.57 & 94.83 & 92.70 & 91.09 & 94.02 & 93.20 & 92.59 & 93.02 & 91.37 & 92.33 & 95.48 & 89.67 & 93.56 & 92.90 & 92.81 \\
\hline
\rowcolor{headergray} \multicolumn{18}{l}{\textit{Proprietary Models (Offline), Single-Turn Evaluation}} \\
Gemini 1.5 Pro~\cite{reid2024gemini} & 1 FPS & 85.91& 66.97& 79.31& 58.43& 63.37& 61.96& \textbf{69.32}& 58.59& 76.35& 52.64& \textbf{62.54}& 35.53& 74.24& 61.67& \textbf{57.15}& \textbf{63.00} \\
GPT-4o~\cite{gpt4o} & 64 & 69.80& 64.22& 71.55& 51.12& 70.30& 59.78& \underline{64.46}& 57.91& 75.68& 48.66& \underline{60.75}& 27.58& 73.21& 59.40& 53.40& \underline{59.54} \\
\hline
\rowcolor{headergray} \multicolumn{18}{l}{\textit{Open-Source Models (Offline), Single-Turn Evaluation}} \\
LLaVA-Video-7B~\cite{zhang2024video} & 64 & 69.13& 58.72& 68.83& 49.44& 74.26& 59.78& 63.52& 56.23& 57.43& 7.53& 40.40& 34.10& 69.95& 60.42& \underline{54.82}& 52.91 \\
LLaVA-OV-7B~\cite{li2024llava} & 64 & 66.44& 57.80& 73.28& 53.37& 71.29& 61.96& 64.02& 54.21& 55.41& 21.51& 43.71& 25.64& 67.09& 58.75& 50.50& 52.74 \\
LLaVA-N-Video-7B~\cite{li2024llavanext} & 64 & 69.80& 59.60& 66.40& 50.60& 72.30& 61.40& 63.30& 51.20& 64.20& 9.70& 41.70& 34.10& 67.60& 60.80& 54.20& 53.10 \\
Qwen2-VL-7B~\cite{wang2024qwen2} & 64 & 60.40& 50.46& 56.03& 47.19& 66.34& 55.43& 55.98& 47.81& 35.48& 56.08& 46.46& 31.66& 65.82& 48.75& 48.74& 50.39 \\
InternVL-V2-8B~\cite{chen2024far} & 64 & 67.11& 60.55& 63.79& 46.07& 68.32& 56.52& 60.39& 48.15& 57.43& 24.73& 43.44& 26.50& 59.14& 54.14& 46.60& 50.15 \\
LongVU-7B~\cite{shen2024longvu} & 1 FPS & 55.70& 49.50& 59.50& 48.30& 68.30& 63.00& 57.40& 43.10& 66.20& 9.10& 39.50& 16.60& 69.00& 60.00& 48.50& 48.50 \\
\hline
\rowcolor{headerblue} \multicolumn{18}{l}{\textit{Open-Source Models (Streaming)}} \\
Flash-VStream-7B~\cite{zhang2025flash} & 1 FPS & 24.16& 29.36& 28.45& 33.71& 25.74& 28.80& 28.37& 39.06& 37.16& 5.91& 27.38& 8.02& 67.25& 60.00& 45.09& 33.61 \\
VideoLLM-Online-8B~\cite{chen2024videollm} & 2 FPS & 8.05& 23.85& 12.07& 14.04& 45.54& 21.20& 20.79& 22.22& 18.80& 12.18& 17.73& - & - & - & - & - \\
VideoLLM-EyeWO~\cite{zhang2025eyes} & 1 FPS & 24.16& 27.52& 31.89& 32.58& 44.55& 35.87& 32.76& 39.06& 38.51& 6.45& 28.00& - & - & - & - & - \\
Dispider~\cite{qian2025dispider} & 1 FPS & 57.72& 49.54& 62.07& 44.94& 61.39& 51.63& 54.55& 48.48& 55.41& 4.30& 36.06& 18.05& 37.36& 48.75& 34.72& 41.78 \\
TimeChat-Online-7B~\cite{yao2025timechat} & 1 FPS & 69.80& 48.60& 64.70& 44.90& 68.30& 55.40& 58.60& 53.90& 62.80& 9.10& 42.00& 32.50& 36.50& 40.00& 36.40& 45.60 \\
StreamAgent-7B~\cite{yang2025streamagent} & 1 FPS & 71.20& 53.20& 63.60& 53.90& 67.30& 58.70& 61.30& 54.80& 58.10& 25.80& 41.70& 35.90& 48.40& 52.00& 45.40& 49.40 \\
QueryStream-7B~\cite{zhang2026querystream} & 1 FPS & 74.50& 47.70& 70.70& 46.60& 71.30& 57.60& 61.40& 54.20& 63.50& 8.60& 42.10& 33.20& 43.10& 40.80& 39.03& 47.51 \\
\hline
\rowcolor{headerblue} \multicolumn{18}{l}{\textit{Offline Backbones $\rightarrow$ Online Inference with STRIDE}} \\
Qwen3-VL-8B~\cite{Qwen3-VL} & 1 FPS & 69.80& 59.60& 73.30& 57.30& 71.30& 58.70& 65.00& 55.60& 63.50& 12.90& 44.00& 37.70& 60.80& 40.40& 46.30& 51.77 \\
\hspace{0.5em}+ Baseline-AR~\cite{wang2025streambridge} & 1 FPS & 73.80 & 65.10 & 73.30 & 62.40 & 70.30 & 71.20 & \underline{69.35} & 54.90 & 66.90 & 17.20 & \underline{46.33} & 29.70 & 56.00 & 42.50 & 42.73 & 52.81 \\ \hdashline
Gemma3-4B~\cite{team2025gemma} & 1 FPS & 65.80& 48.60& 56.00& 36.00& 66.30& 50.00& 53.78& 44.40& 41.90& 3.20& 29.83& 14.40& 61.40& 52.50& 42.77& 42.13 \\
\hspace{0.5em}+ \textbf{\textit{STRIDE}} & 1 FPS & 73.20& 60.60& 64.70& 39.30& 71.30& 56.50& 60.93& 47.80& 52.00& 4.80& 34.87& 42.60& 64.60& 60.00& 55.73& 50.51 \gain{8.38} \\
InternVL3-8B~\cite{wang2025internvl3} & 1 FPS & 65.80& 52.30& 68.10& 51.10& 71.30& 62.00& 61.77& 58.90& 66.90& 9.70& 45.17& 36.60& 64.10& 43.30& 48.00& 51.64 \\
\hspace{0.5em}+ \textbf{\textit{STRIDE}} & 1 FPS & 75.80& 54.10& 80.20& 56.70& 74.30& 65.20& 67.72& 58.90& 65.50& 11.30& 45.23& 40.10& 67.70& 66.20& \underline{58.00}& \underline{56.98} \gain{5.34} \\
Qwen3-VL-8B~\cite{Qwen3-VL} & 1 FPS & 69.80& 59.60& 73.30& 57.30& 71.30& 58.70& 65.00& 55.60& 63.50& 12.90& 44.00& 37.70& 60.80& 40.40& 46.30& 51.77 \\
\hspace{0.5em}+ \textbf{\textit{STRIDE}} & 1 FPS & 76.50& 64.20& 79.30& 61.20& 73.30& 63.60& \textbf{69.68}& 57.20& 72.30& 14.00& \textbf{47.83}& 46.40& 63.10& 69.60& \textbf{59.70}& \textbf{59.07} \gain{7.30} \\
\Xhline{2\arrayrulewidth}
\end{tabular}
}
\vspace{-3mm}
\end{table*}


\subsection{Qualitative Results on Streaming Video Understanding}
\label{subsec:main_results_realtime}
\definecolor{headerblue}{HTML}{ECF4FF}
\definecolor{headergray}{HTML}{F2F2F2}
\definecolor{deltagreen}{HTML}{7CC472}

\begin{table*}[t]
\centering
\small
\setlength{\tabcolsep}{3.5pt}
\caption{Evaluation results on StreamingBench~\cite{lin2024streamingbench}. Baseline-AR uses autoregressive binary prediction. Offline models follow the original single-turn protocol with segmented clips, whereas streaming methods process frames sequentially.}
\label{tab:streaming_bench_full}
\resizebox{1.0\linewidth}{!}{
    \begin{tabular}{@{}l c cccccccccc c !{\vrule} cccc c !{\vrule} cccc c !{\vrule} c@{}}
\Xhline{2\arrayrulewidth}
\multirow{2}{*}[-4pt]{Method} & \multirow{2}{*}[-4pt]{\shortstack{\# of\\Frames}} & \multicolumn{11}{c}{\textbf{Real-Time Visual Understanding}} & \multicolumn{5}{c}{\textbf{Omni-Source Understanding}} & \multicolumn{5}{c}{\textbf{Contextual Understanding}} & \textbf{Overall} \\
\cmidrule(l{2pt}r{2pt}){3-13} \cmidrule(l{2pt}r{2pt}){14-18} \cmidrule(l{2pt}r{2pt}){19-23} \cmidrule(l{2pt}r{2pt}){24-24}
& & OP & CR & CS & ATP & EU & TR & PR & SU & ACP & CT & Avg. & ER & SCU & SD & MA & Avg. & ACU & MCU & SQA & PO & Avg. & Avg. \\
\hline
\rowcolor{headergray} \multicolumn{24}{l}{\textit{Human}} \\
Human & - & 89.47 & 92.00 & 93.60 & 91.47 & 95.65 & 92.52 & 88.00 & 88.75 & 89.74 & 91.30 & 91.46 & 88.00 & 88.24 & 93.60 & 90.27 & 90.26 & 88.80 & 90.40 & 95.00 & 100 & 93.55 & 91.66 \\
\hline
\rowcolor{headergray} \multicolumn{24}{l}{\textit{Proprietary Models (Offline)}} \\
Gemini 1.5 pro~\cite{reid2024gemini} & 1 FPS & 79.02& 80.47& 83.54& 79.67& 80.00& 84.74& 77.78& 64.23& 71.95& 48.70& \textbf{75.69}& 46.80& 39.60& 74.90& 80.00& \textbf{60.22}& 51.41& 40.73& 54.80& 45.10& \textbf{48.73}& \textbf{67.07} \\
GPT-4o~\cite{gpt4o} & 64 & 77.11& 80.47& 83.91& 76.47& 70.19& 83.80& 66.67& 62.19& 69.12& 49.22& \underline{73.28}& 41.20& 37.20& 43.60& 56.00& \underline{44.50}& 41.20& 38.40& 32.80& 56.86& \underline{38.70}& \underline{60.15} \\
\hline
\rowcolor{headergray} \multicolumn{24}{l}{\textit{Open-Source Models (Offline)}} \\
LLaVA-OV-7B~\cite{li2024llava} & 32 & 80.38& 74.22& 76.03& 80.72& 72.67& 71.65& 67.59& 65.45& 65.72& 45.08& 71.12& 40.80& 37.20& 33.60& 44.80& 38.40& 35.60& 36.00& 27.27& 29.55& 32.74& 56.36 \\
Qwen2-VL-7B~\cite{wang2024qwen2} & 1 FPS & 75.20& 82.81& 73.19& 77.45& 68.32& 71.03& 72.22& 61.19& 61.47& 46.11& 69.04& 41.20& 22.00& 32.80& 43.60& 34.90& 31.20& 26.00& 39.60& 22.73& 31.66& 54.14 \\
MiniCPM-V 2.6 8B~\cite{yao2024minicpm} & 32 & 71.93& 71.09& 77.92& 75.82& 64.60& 65.73& 70.37& 56.10& 62.32& 53.37& 67.44& 40.80& 24.00& 34.00& 41.20& 35.00& 34.00& 31.60& 41.92& 22.22& 34.97& 53.85 \\
InternVL-V2-8B~\cite{chen2024far} & 16 & 68.12& 60.94& 69.40& 77.12& 67.70& 62.93& 59.26& 53.25& 54.96& 56.48& 63.72& 37.60& 26.40& 37.20& 42.00& 35.80& 32.00& 31.20& 32.32& 40.91& 32.42& 51.40 \\
Kangaroo-7B~\cite{liu2024kangaroo} & 64 & 71.12& 84.38& 70.66& 73.20& 67.08& 61.68& 56.48& 55.69& 62.04& 38.86& 64.60& 37.60& 31.20& 28.80& 39.20& 34.20& 32.80& 26.40& 33.84& 16.00& 30.06& 51.10 \\
LongVA-7B~\cite{zhang2024long} & 128 & 70.03& 63.28& 61.20& 70.92& 62.73& 59.50& 61.11& 53.66& 54.67& 34.72& 59.96& 39.60& 32.40& 28.00& 41.60& 35.40& 32.80& 29.60& 30.30& 15.91& 29.95& 48.66 \\
VILA-1.5-8B~\cite{lin2024vila} & 14 & 53.68& 49.22& 70.98& 56.86& 53.42& 53.89& 54.63& 48.78& 50.14& 17.62& 52.32& 41.60& 26.40& 28.40& 36.00& 33.10& 26.80& 34.00& 23.23& 17.65& 27.35& 43.20 \\
Video-LLaMA2-7B~\cite{cheng2024videollama} & 32 & 55.86& 55.47& 57.41& 58.17& 52.80& 43.61& 39.81& 42.68& 45.61& 35.23& 49.52& 30.40& 32.40& 30.40& 36.00& 32.40& 24.80& 26.80& 18.67& 0.00& 21.93& 40.40 \\
\hline
\rowcolor{headerblue} \multicolumn{24}{l}{\textit{Open-Source Models (Streaming)}} \\
Flash-VStream-7B~\cite{zhang2025flash} & 1 FPS & 25.89& 43.57& 24.91& 23.87& 27.33& 13.08& 18.52& 25.20& 23.87& 48.70& 23.23& 25.91& 24.90& 25.60& 28.40& 26.00& 24.80& 25.20& 26.80& 1.96& 24.12& 24.04 \\
VideoLLM-Online-8B~\cite{chen2024videollm} & 2 FPS & 39.07& 40.06& 34.49& 31.05& 45.54& 32.40& 31.48& 34.16& 42.49& 27.89& 35.99& 31.20& 26.51& 24.10& 32.00& 28.45& 24.19& 29.20& 30.80& 3.92& 26.55& 32.48 \\
Dispider~\cite{qian2025dispider} & 1 FPS & 74.92& 75.53& 74.10& 73.08& 74.44& 59.92& 76.14& 62.91& 62.16& 45.80& 67.63& 35.46& 25.26& 38.57& 43.34& 35.66& 39.62& 27.65& 34.80& 25.34& 33.61& 53.12 \\
StreamAgent~\cite{yang2025streamagent} & 1 FPS & 79.63& 78.31& 79.28& 75.87& 74.74& 76.92& 82.94& 66.31& 73.69& 55.40& \underline{74.31}& 35.86& 26.26& 38.87& 44.04& 36.26& 39.72& 30.25& 39.60& 28.90& 34.62& 57.02 \\
TimeChat-Online-7B~\cite{yao2025timechat} & 1 FPS & 80.80& 79.70& 80.80& 83.30& 74.80& 78.80& 78.70& 64.20& 68.80& 58.00& \textbf{75.28}& - & - & - & - & - & - & - & - & - & - & - \\
QueryStream-7B~\cite{zhang2026querystream} & 1 FPS & 82.11& 83.59& 78.23& 82.69& 75.47& 80.06& 79.63& 63.01& 67.90& 42.55& 74.04& - & - & - & - & - & - & - & - & - & - & - \\
\hline
\rowcolor{headerblue} \multicolumn{24}{l}{\textit{Offline Backbones $\rightarrow$ Online Inference with STRIDE}} \\
Qwen3-VL-8B~\cite{Qwen3-VL} & 1 FPS & 62.70& 68.00& 69.70& 53.30& 67.50& 65.10& 67.60& 48.00& 68.00& 40.90& 60.88& 36.80& 18.00& 32.80& 34.00& 30.40& 25.60& 23.60& 31.20& 32.40& 28.20& 46.84 \\
\hspace{0.5em}+ Baseline-AR~\cite{wang2025streambridge} & 1 FPS & 79.00& 72.70& 85.50& 70.80& 73.30& 76.60& 81.50& 63.40& 80.70& 44.60& 73.79& 45.20& 29.20& 35.20& 35.20& 36.20& 35.60& 37.20& 48.40& 24.30& 36.38& 57.12 \\ \hdashline
Gemma3-4B~\cite{team2025gemma} & 1 FPS & 63.80& 69.50& 68.80& 54.40& 60.20& 65.10& 59.30& 40.70& 62.70& 21.80& 57.49& 28.80& 31.20& 30.40& 46.40& 34.20& 34.40& 31.20& 38.80& 12.40& 29.20& 46.03 \\
\hspace{0.5em}+ \textbf{\textit{STRIDE}} & 1 FPS & 66.80& 71.90& 66.60& 57.20& 66.50& 70.70& 60.20& 43.10& 65.00& 23.80& 60.00& 35.60& 31.60& 36.00& 44.00& 36.80& 33.60& 35.20& 44.80& 41.60& \underline{38.80}& 50.14 \\
InternVL3-8B~\cite{wang2025internvl3} & 1 FPS & 74.90& 82.00& 75.70& 61.20& 72.00& 67.60& 74.10& 66.70& 78.10& 34.20& 68.71& 40.40& 27.60& 38.80& 45.60& 38.10& 38.00& 26.00& 36.80& 31.20& 33.00& 53.97 \\
\hspace{0.5em}+ \textbf{\textit{STRIDE}} & 1 FPS & 74.90& 78.90& 76.70& 68.60& 77.00& 77.30& 77.80& 71.50& 83.00& 33.20& 72.45& 39.60& 22.40& 44.00& 50.80& \underline{39.20}& 34.00& 35.20& 43.20& 42.80& \underline{38.80}& \underline{57.58}  \\
Qwen3-VL-8B~\cite{Qwen3-VL} & 1 FPS & 62.70& 68.00& 69.70& 53.30& 67.50& 65.10& 67.60& 48.00& 68.00& 40.90& 60.88& 36.80& 18.00& 32.80& 34.00& 30.40& 25.60& 23.60& 31.20& 32.40& 28.20& 46.84 \\
\hspace{0.5em}+ \textbf{\textit{STRIDE}} & 1 FPS & 77.10& 75.00& 77.30& 72.80& 76.40& 77.90& 76.90& 69.90& 84.30& 46.10& 74.24& 42.80& 24.00& 45.20& 53.20& \textbf{41.30}& 32.00& 38.40& 46.40& 42.80& \textbf{39.90}& \textbf{59.29} \\
\Xhline{2\arrayrulewidth}
\end{tabular}
}
\end{table*}
\cref{tab:ovo_bench_full,tab:streaming_bench_full} present results on OVO-Bench and StreamingBench, respectively. The proposed \textit{STRIDE} outperforms the autoregressive baseline (\ie, Baseline-AR) \cite{wang2025streambridge} by introducing the proposed masked denoising process. Furthermore, across all three downstream models~\cite{team2025gemma,wang2025internvl3,Qwen3-VL} on OVO-Bench,
\textit{STRIDE} achieves significant gains in Forward Active Responding, which directly evaluates proactive when-to-speak control. This setting benefits from our span-structured prediction, which models response timing over temporal activation region rather than through independent per-frame decisions. \textit{STRIDE} also consistently improves Real-Time Visual Perception, indicating that stable triggering allows the downstream Video-LLM to ingest well-scoped visual context at the appropriate moment.  On StreamingBench, this advantage extends broadly across all three evaluation dimensions: Real-time Visual Understanding, Omni-Source Understanding, and Contextual Understanding, with the most notable improvements in Proactive Output (PO) subtask that requires the model to determine response timing without explicit timing cue. Together, these results suggest that the proposed framework reliably enhances both the precision of when to respond and the relevance of responses across diverse streaming conditions.

\subsection{Activation Evaluation via Temporal Grounding}
\label{subsec:proactive_activation}
Accurate temporal grounding is central to proactive activation. While the previous benchmarks~\cite{niu2025ovo,lin2024streamingbench} evaluate the end-to-end behavior of streaming pipelines, they do not directly measure the quality of the activation model itself. To isolate this component, we evaluate the activation model independently on ET-Bench~\cite{liu2024bench}, which focuses on fine-grained event-level temporal understanding. As shown in \cref{tab:results_activate}, the gain from replacing binary classification with masked diffusion is substantial: \textit{STRIDE} outperforms Baseline-AR by 27.1 on TVG and 8.3 on average, demonstrating that structured sequence denoising provides considerably sharper boundary resolution than per-frame supervision. Notably, \textit{STRIDE} achieves these results with only 2B parameters, outperforming both streaming baselines and temporal-localization specialized MLLMs of standard size (7--13B parameters) on the overall average.

\subsection{Ablation Studies on STRIDE Components}
\subsubsection{Effects of Masking Strategies.}
\cref{tab:ablation}(a) evaluates how different masking strategies affect the learning of span-structured activation sequences. The standard MDM protocol independent masking performs the worst across all metrics, indicating that activation prediction cannot be treated as independent point-wise denoising since it fails to capture the temporal structure of activation transitions. To better reflect the span-level nature of activation, we adopt three complementary masking patterns described in \cref{sec:training}: boundary-anchored span masking (Span), full masking (Full), and span unmasking ($\overline{\text{Span}}$). Span masks contiguous regions near activation boundaries, Full masks the entire sequence to simulate the cold-start condition, and $\overline{\text{Span}}$ exposes boundary refinement patterns encountered during denoising. Combining these strategies significantly improves performance, suggesting that diverse span corruption patterns help the model learn coherent activation spans for better boundary prediction.

\begin{table*}[t]
\centering
\begin{minipage}[t]{0.56\linewidth}
  \centering
\small
\caption{Online activation accuracy on ET-Bench~\cite{liu2024bench}. The Baseline-AR uses autoregressive prediction, while other setups are the same as \textit{STRIDE}.}
\label{tab:results_activate}
\vspace{-2mm}
\resizebox{0.99\linewidth}{!}{
\begin{tabular}{@{}l c c ccccc c@{}}
\Xhline{2\arrayrulewidth}
& Frames & Params & TVG$_{\text{F1}}$ & EPM$_{\text{F1}}$ & TAL$_{\text{F1}}$ & DVC$_{\text{F1}}$ & SLC$_{\text{F1}}$ & Avg \\
\hline
\rowcolor{headerblue}
\multicolumn{9}{l}{\textit{Temporal-Localization Specialized MLLMs}}\\
VTimeLLM~\cite{huang2024vtimellm} & 100 & 7B & 7.6 & 1.9 & 18.2 & 12.4 & 8.7 & 9.8 \\
VTG-LLM~\cite{guo2025vtg} & 96 & 7B & 15.9 & 3.7 & 14.4 & 40.2 & 20.8 & 19.0\\
TimeChat~\cite{ren2024timechat} & 96 & 7B & 26.2 & 3.9 & 10.1 & 16.6 & 5.6 & 12.5\\
LITA~\cite{huang2024lita} & 100 & 13B & 22.2 & 4.6 & 18.0 & 39.7 & 21.0 & 21.1\\
ETChat~\cite{liu2024bench} & 1 FPS & 5B & 38.6 & 10.2 & 30.8 & 38.4 & 24.4 & 28.5\\
\rowcolor{headerblue}
\multicolumn{9}{l}{\textit{Streaming Baselines}}\\
VideoLLM-Online~\cite{chen2024videollm} & 2 FPS & 8B & 13.2 & 3.8 & 9.1 & 24.0 & 9.9 & 12.0\\
Dispider~\cite{qian2025dispider} & 1 FPS & 9B & \underline{36.1} & \textbf{15.5} & \textbf{27.3} & 33.8 & 18.8 & \underline{26.3}\\
StreamBridge~\cite{wang2025streambridge} & 1 FPS & 8B & 34.3 & -- & 24.3 & \underline{38.3} & \underline{22.6} & -- 
\\[2pt] \cdashline{1-9}[6pt/4pt] \noalign{\vskip 3pt}
\textbf{Baseline-AR}~\cite{wang2025streambridge} & 1 FPS & 2B & 35.7 & 2.5 & 21.2 & \textbf{39.6} & \underline{22.6} & 24.3\\
\textbf{\textit{STRIDE}} & 1 FPS & 2B & \textbf{62.8} & \underline{10.7} & \underline{24.6} & 36.5 & \textbf{28.5} & \textbf{32.6}\\
\Xhline{2\arrayrulewidth}
\end{tabular}
}
\end{minipage}
\hfill
\begin{minipage}[t]{0.43\linewidth}
  \centering
\small
\caption{Ablation studies on ET-Bench evaluating (a) masking strategy design, (b) sequence duplication, and (c) selective re-masking.}
\label{tab:ablation}
\resizebox{0.99\linewidth}{!}{
\begin{tabular}{@{}l ccccc c@{}}
\Xhline{2\arrayrulewidth}
& TVG$_{\text{F1}}$ & EPM$_{\text{F1}}$ & TAL$_{\text{F1}}$ & DVC$_{\text{F1}}$ & SLC$_{\text{F1}}$ & Avg \\
\hline
\rowcolor[HTML]{ECF4FF} \multicolumn{7}{l}{\textit{(a) Masking Strategy}} \\
Indep.\ only                           & 8.5 & 3.3 & 6.1 & 8.8 & 9.2 & 7.2 \\
Span only                              & 30.6 & 6.1 & 22.9 & 25.4 & 20.6 & 21.1 \\
Span + Full                            & 36.8 & 7.0 & \textbf{26.0} & 24.0 & 21.3 & 23.0 \\
Span + Full + $\overline{\text{Span}}$ & \textbf{62.8} & \textbf{10.7} & 24.6 & \textbf{36.5} & \textbf{28.5} & \textbf{32.6} \\
\hline
\rowcolor[HTML]{ECF4FF} \multicolumn{7}{l}{\textit{(b) Sequence Duplication}} \\
w/o Seq.\ Duplication                  & 49.6 & 6.0 & 23.6 & 19.9 & 15.2 & 22.9 \\
w/\ \ Seq.\ Duplication & \textbf{62.8} & \textbf{10.7} & \textbf{24.6} & \textbf{36.5} & \textbf{28.5} & \textbf{32.6} \\
\hline
\rowcolor[HTML]{ECF4FF} \multicolumn{7}{l}{\textit{(c) Selective Re-masking}} \\
w/o\ \ Re-masking (last-only) & 39.5 & 2.5 & 19.1 & 30.7 & 21.2 & 22.6 \\
w/\ \ Re-masking (selective) & \textbf{62.8} & \textbf{10.7} & \textbf{24.6} & \textbf{36.5} & \textbf{28.5} & \textbf{32.6} \\
\Xhline{2\arrayrulewidth}
\end{tabular}
}
 
\end{minipage}
\end{table*}


\begin{figure*}[t]
\centering
\includegraphics[width=0.99\linewidth]{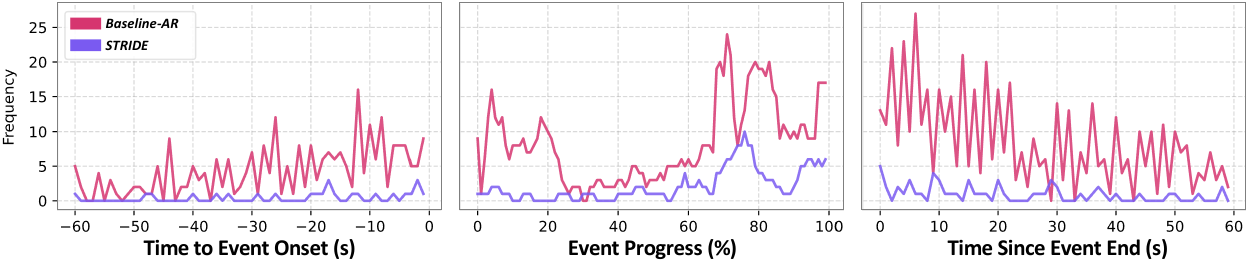}
\begin{flushleft}
\small
    {\hspace{2.5cm}(a) Pre-Event \hspace{3.3cm}(b) During Event\hspace{3.3cm}(c) Post-Event}
\end{flushleft}	
\caption{Activation transition frequency results around event boundaries on ET-Bench TVG. Baseline-AR model shows frequent oscillations near boundaries, whereas \textit{STRIDE} produces more robust activation spans.}
\label{fig:flickering}
\end{figure*}

\subsubsection{Effect of Sequence Duplication.}
Masked diffusion relies on contextual reasoning over the activation window, whereas the pretrained backbone used in \textit{STRIDE} follows causal attention and therefore only exposes left context during prediction. This mismatch limits the model’s ability to jointly infer activation states within the window. To mitigate this, we apply \textit{sequence duplication}, which provides full-window context to the prediction tokens while preserving the causal backbone. As shown in \cref{tab:ablation}(b), removing sequence duplication leads to a consistent performance drop across all tasks, reducing the average score from 32.6 to 22.9. The degradation is particularly notable in temporally sensitive tasks such as TVG and DVC, indicating that accurate boundary reasoning benefits from access to the full activation window. These results demonstrate that sequence duplication effectively recovers bidirectional context for diffusion-based refinement, enabling full-window conditioning through a simple input reparameterization without modifying the causal architecture.

\subsubsection{Effect of Selective Re-masking.}
In the streaming setting, activation predictions are carried forward as the window advances. If these states are preserved without revision, early mistakes can propagate and gradually corrupt the activation sequence. To examine this effect, we compare our selective re-masking strategy with a simplified variant that masks only the newly appended position (\textit{last-only}), leaving previous decisions fixed. As shown in \cref{tab:ablation}(c), restricting re-masking to the last position leads to a substantial performance drop, reducing the average score from 32.6 to 22.6. As only predicting last token falls into autoregressive prediction, the resulting performance is also similar to \textit{Baseline-AR} in \cref{tab:results_activate}. In contrast, selectively re-masking low-confidence positions allows the model to revise uncertain decisions as new frames arrive, enabling refinement of the activation sequence by using the updated context information.

\subsection{Behavioral Analysis of STRIDE Properties}
\subsubsection{Flickering Analysis around Event Boundaries.}
While ET-Bench quantifies activation accuracy, it is also limited to capture the temporal stability of activation decisions. In particular, per-frame activation models may suffer from flickering behavior due to their inherently isolated predictions, where predictions rapidly oscillate between active and inactive states (0$\leftrightarrow$1), resulting in unstable triggering and poorly resolved transition boundaries. To analyze this phenomenon, we measure the frequency of activation transitions relative to event boundaries. Specifically, we align predictions around ground-truth events and accumulate transition counts within three regions (\cref{fig:flickering}): (a) pre-event (-60s to onset), (b) during-event (normalized by event progress \%), and (c) post-event (offset to +60s), using the TVG task of ET-Bench where each instance corresponds to a single event.

As in the figure, across all regions, \textit{Baseline-AR} exhibits substantially higher transition frequency, indicating unstable activation sequences with frequent on/off oscillations. This instability becomes particularly striking near event boundaries, where transition frequency sharply increases, suggesting difficulty in resolving the precise onset and offset of events. In contrast, \textit{STRIDE} produces significantly smoother activation patterns with far fewer transitions. The reduced flickering indicates that modeling activation as structured sequence denoising encourages temporally coherent predictions, allowing the model to maintain consistent activation spans and more reliably capture event boundaries.

\subsubsection{Latency--Accuracy Trade-off for Denoising Steps.}
\begin{figure}[t]
  \centering
  \begin{minipage}[c]{0.54\linewidth}
    \centering
      \includegraphics[width=\linewidth]{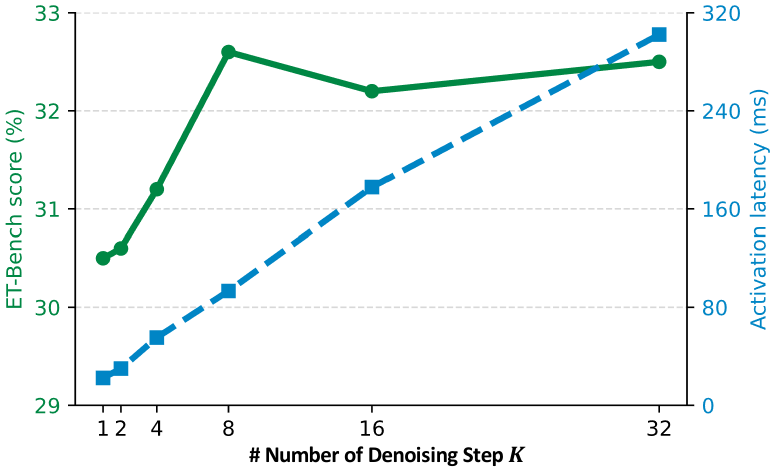}
      \captionof{figure}{Trade-off between ET-Bench performance (mean F1) and inference latency for denoising step $K$.}
      \label{fig:ablation_sampling}
  \end{minipage}
  \hfill
  \begin{minipage}[c]{0.45\linewidth}
    \centering
\small
\captionof{table}{Latency and VRAM usage of the downstream Video-LLM and \textit{STRIDE} activation modules (with AR variation) during streaming inference.}
\label{tab:resource}
\resizebox{\linewidth}{!}{
\begin{tabular}{@{}lcc@{}}
\Xhline{2\arrayrulewidth}
\textbf{Procedure} & \textbf{Latency (ms)} & \textbf{VRAM} \\
\hline
\rowcolor[HTML]{ECF4FF} \multicolumn{3}{l}{\textit{Downstream Video-LLM}} \\
Response Gen. (TTFT) & 1276 & 17.8\,GB \\
Response Gen. (TTLT)  &  1511 & + 13\,MB \\
\rowcolor[HTML]{ECF4FF} \multicolumn{3}{l}{\textit{STRIDE}} \\
Activation Sate (Base) &  & 5.2\,GB \\
+ 1 Denoising Step  &  12 & + 10\,MB \\
+ Append Frame              &  20 & + 30\,MB \\
\rowcolor[HTML]{ECF4FF} \multicolumn{3}{l}{\textit{Baseline-AR}} \\
Activation Sate (Base) &  & 5.2\,GB \\
+ Append Frame              &  26 & + 30\,MB \\
\Xhline{2\arrayrulewidth}
\end{tabular}
}
  
    \label{tab:resource}
  \end{minipage}
\end{figure}

We analyze the effect of the denoising step $K$ on both activation model accuracy and inference latency as illustrated in~\cref{fig:ablation_sampling}. This trade-off is particularly important in the streaming setting, where the activation model operates online and directly determines the model’s response latency. Increasing $K$ allows the model to perform more refinement steps, improving activation accuracy but also increasing inference time. In practice, we observe that performance saturates quickly: around $K=8$ steps already achieves near-maximum mean F1 across ET-Bench subtasks. This behavior likely stems from the small output space of the activation sequence, where each position only takes binary states (0 or 1), making the denoising process relatively easier to converge than large vocabulary space. At $K=8$, the inference latency is approximately $\sim$100\,ms, which is practical enough to support real-time operation for streaming frame rates of downstream models.

\subsubsection{Streaming Efficiency and Memory Footprint.}
Extending the latency-accuracy trade-off analysis in \cref{fig:ablation_sampling}, we decompose the computational overhead introduced by the activation model under a streaming setup. The measurement is conducted on a single H100 GPU with a 128-frame context budget. As shown in \cref{tab:resource}, when a subsequent response is required, the 113 ms (new frame and $K\!=\!8$ denoising steps) added by \textit{STRIDE} incurs only a 7\% additional latency compared to the 1511 ms required by the base model Qwen3VL-8B~\cite{Qwen3-VL} without the triggering module. When a trigger is unnecessary, \textit{STRIDE} saves approximately 91\% of the total processing time (113 ms vs. 1276 ms). Furthermore, compared to the per-frame decision baseline (Baseline-AR, 26 ms), the extra latency from the diffusion process (113 ms) is limited to 87 ms. In terms of memory, \textit{STRIDE} maintains a lightweight footprint of 5.2 GB. Executing denoising process requires an additional 10 MB, and each new frame introduces 30 MB of incremental memory usage. These highlight the advantage of the two-stage design: a lightweight activation model gates the expensive downstream model. Even with the masked diffusion module employed in STRIDE, trigger modeling introduces only minimal latency and memory overhead, maintaining efficient streaming inference.

\section{Conclusion}
We present \textit{STRIDE}, a framework for proactive streaming video understanding that models activation as a structured temporal sequence rather than independent per-frame decisions. By leveraging a lightweight masked diffusion module to jointly refine activation signals over a sliding window, \textit{STRIDE} captures span-level temporal structure and produces more stable and coherent triggering behavior in streaming settings. Extensive experiments and analyses show that jointly modeling activation over a temporal window significantly improves event boundary localization and reduces unstable triggering, while introducing only minimal overhead to the overall streaming pipeline.

\bibliographystyle{plainnat}
\bibliography{citation}

\clearpage
\newpage

\appendix
\definecolor{qaanswer}{RGB}{57,255,20}
\definecolor{qawrong}{RGB}{214,39,40}

\newcommand{\ANS}[1]{\textcolor{qaanswer}{#1}}
\newcommand{\WA}[1]{\textcolor{qawrong}{#1}}

\def\framevscale{}
\newcommand{\FrameCell}[2]{%
    \begin{minipage}[t]{0.235\linewidth}
        \centering
        \ifx\framevscale\empty
            \includegraphics[width=\linewidth]{#1/frame#2.jpg}\\
        \else
            \includegraphics[width=\linewidth,height=\framevscale\linewidth]{#1/frame#2.jpg}\\
        \fi
        \vspace{-7pt}
        {\scriptsize \input{#1/time#2.tex}}
        \vspace{1pt}
    \end{minipage}
}

\newcommand{\RenderSample}[1]{%

    \vspace{-7pt}
    \begin{center}
        \includegraphics[width=\linewidth]{#1/timebar.pdf}
    \end{center}

    \begin{center}
    \setlength{\tabcolsep}{0.0025\linewidth}
    \renewcommand{\arraystretch}{1.2}
    \begin{tabular}{c c c c c c c c}
        \input{#1/time1.tex} &
        \input{#1/time2.tex} &
        \input{#1/time3.tex} &
        \input{#1/time4.tex} \\
        \input{#1/time5.tex} &
        \input{#1/time6.tex} &
        \input{#1/time7.tex} &
        \input{#1/time8.tex} \\
        \input{#1/time9.tex} &
        \input{#1/time10.tex} &
        \input{#1/time11.tex} &
        \input{#1/time12.tex} \\
        \input{#1/time13.tex} &
        \input{#1/time14.tex} &
        \input{#1/time15.tex} &
        \input{#1/time16.tex} \\
        \input{#1/time17.tex} &
        \input{#1/time18.tex} &
        \input{#1/time19.tex} &
        \input{#1/time20.tex} \\
        \input{#1/time21.tex} &
        \input{#1/time22.tex} &
        \input{#1/time23.tex} &
        \input{#1/time24.tex} \\
        \input{#1/time25.tex} &
        \input{#1/time26.tex} &
        \input{#1/time27.tex} &
        \input{#1/time28.tex} \\
        \input{#1/time29.tex} &
        \input{#1/time30.tex} &
        \input{#1/time31.tex} &
        \input{#1/time32.tex}
    \end{tabular}
    \end{center}
    \vspace{-10pt}
}

\newcommand{\RenderQAExample}[1]{%

    \begin{tcolorbox}[colback=gray!10,colframe=black]
    \normalsize
    \textbf{Question:}\\
    \vspace{1pt}
    {\small \input{#1/question.tex}}

    \vspace{2pt}
    \textbf{Options:}
    {\small
    \begin{itemize}[nosep]
        \input{#1/options.tex}
    \end{itemize}
    }

    \end{tcolorbox}
}

\newcommand{\RenderETExample}[1]{%

    \begin{tcolorbox}[colback=gray!10,colframe=black]
    \scriptsize
    \textbf{Question:}\\
    \vspace{2pt}
    \input{#1/question.tex}

    \vspace{-2pt}
    \end{tcolorbox}
}

\newcommand{\RenderQAExampleFigure}[3][]{%

    \begin{figure*}[h!]
        \centering
        \begin{minipage}{0.75\linewidth}
            %

    \begin{tcolorbox}[colback=gray!10,colframe=black]
    \normalsize
    \textbf{Question:}\\
    \vspace{1pt}
    {\small \input{#2/question.tex}}

    \vspace{2pt}
    \textbf{Options:}
    {\small
    \begin{itemize}[nosep]
        \input{#2/options.tex}
    \end{itemize}
    }

    \end{tcolorbox}

            \def\framevscale{#1}%
            \RenderSample{#2}
            \def\framevscale{}%
        \end{minipage}
        \vspace{4pt}
        \caption{#3}
        \label{fig:qual-#2}
    \end{figure*}
}

\newcommand{\RenderETExampleFigure}[3][]{%

    \begin{figure*}[h!]
        \centering
        \begin{minipage}{0.75\linewidth}
            %

    \begin{tcolorbox}[colback=gray!10,colframe=black]
    \scriptsize
    \textbf{Question:}\\
    \vspace{2pt}
    \input{#2/question.tex}

    \vspace{-2pt}
    \end{tcolorbox}

            \def\framevscale{#1}%
            \RenderSample{#2}
            \def\framevscale{}%
        \end{minipage}
        \caption{#3}
        \label{fig:qual-#2}
    \end{figure*}
}

\clearpage
\appendix

\section*{Appendix Contents}
\startcontents[appendices]
\printcontents[appendices]{}{1}{}

\section{Detailed Training Setup for STRIDE}
\subsection{Training Dataset Configuration}
\label{sec:appendix_dataset}
We build the training data for activation span modeling by collecting and carefully curating seven publicly available video understanding datasets~\cite{caba2015activitynet,liu2024bench,sigurdsson2016hollywood, sigurdsson2018charades,wang2024grounded,anne2017didemo,liu2024bench} that provide temporal annotations of events or actions. These datasets cover tasks such as dense video captioning, temporal activity localization, grounded video question answering, and procedural understanding. 
For each dataset, we use the provided temporal boundaries of events to construct activation spans that indicate when a relevant event occurs in the video.

To make the data suitable for our objective, we reorganize the annotations into a unified format where each training sample consists of a video, a query describing the event of interest, and the corresponding temporal span defined by the event start and end times. 
Based on this span, we construct an activation signal over the video timeline: frames (or tokens) that fall within the annotated event span are labeled as \texttt{1} (\textit{active}), while all other positions are labeled as \texttt{0} (\textit{inactive}). 
This binary activation sequence serves as the supervision signal for training the model to detect when the queried event becomes relevant in the video stream.

\begin{table}[h!]
\centering
\small
\setlength{\tabcolsep}{8pt}
\caption{Training dataset statistics. \textit{Videos}: number of source videos; \textit{Items}: total annotation entries; \textit{Single}: single-event entries; \textit{Multi}: multi-event entries (average event count in parentheses); \textit{Segs}: total training segments.}
\label{tab:dataset_config}
\resizebox{0.99\linewidth}{!}{
\begin{tabular}{@{}l rrrrr@{}}
\Xhline{2\arrayrulewidth}
Dataset & Videos & Items & Single & Multi & Segs \\
\hline
ActivityNet-Captions~\cite{caba2015activitynet} & 10,009 & 37,421 & 37,421 & -- & 37,421 \\
LITA~\cite{huang2024lita}                 & 10,000 & 32,489 & 32,489 & -- & 32,489 \\
YouCook2~\cite{zhou2018towards}             & 1,333  & 11,480 & 10,337 & 1,143 ($\times$7.7) & 19,174 \\
ET-Instruct~\cite{liu2024bench}          & 91,121 & 136,072 & 71,966 & 64,106 ($\times$5.1) & 398,609 \\
Charades~\cite{sigurdsson2016hollywood}             & 7,811  & 48,684 & 48,067 & 617 ($\times$2.1) & 49,374 \\
CharadesEgo~\cite{sigurdsson2018charades}          & 6,158  & 61,575 & 57,828 & 3,747 ($\times$2.0) & 65,488 \\
DiDeMo~\cite{anne2017didemo}               & 8,208  & 22,911 & 22,911 & -- & 22,911 \\
Grounded-VideoLLM~\cite{wang2024grounded}    & 17,096 & 61,812 & 61,812 & -- & 61,812 \\
\hline
\textbf{Total}       & \textbf{151,736} & \textbf{412,444} & \textbf{342,831} & \textbf{69,613} & \textbf{687,278} \\
\Xhline{2\arrayrulewidth}
\end{tabular}
}
\end{table}

\cref{tab:dataset_config} summarizes the statistics of the curated training datasets. 
Training samples are constructed differently depending on whether a query corresponds to a single event or multiple events in the video. 
For datasets such as dense video captioning~\cite{caba2015activitynet,liu2024bench}, temporal activity detection~\cite{sigurdsson2016hollywood,sigurdsson2018charades}, grounded video QA~\cite{wang2024grounded}, and moment localization~\cite{anne2017didemo}, each caption or query typically describes a single event. In these cases, the query is paired with the corresponding video segment, and the activation span is defined using the annotated start and end timestamps of the event. 
For datasets involving multiple actions or procedural steps~\cite{sigurdsson2016hollywood,sigurdsson2018charades,zhou2018towards,liu2024bench}, a single query may correspond to multiple events in the video. For action recognition tasks, the action label itself is used as the query, while for procedural datasets we use the original question or caption as the query and construct activation spans for each relevant event. To prevent repeated activation for events that have already occurred, we sample a random time point between the end of the previous event and the start of the current event and set all activation positions before that point to \texttt{0}. This encourages the model to ignore previously completed events and focus on the target span.

Overall, the training set contains 141.7K videos with 379.9K annotations, including 310.3K single-event and 69.6K multi-event samples, resulting in 654.7K training segments. For multi-event samples, the average number of events per sample is reported in parentheses.

\subsection{Training Hyperparameters}
\label{sec:appendix_hparams}
To ensure reproducibility, we report the full set of hyperparameters used for training \textit{STRIDE} (Qwen3-VL 2B and 4B) in \Cref{tab:train_hparams}.
We process the input video stream at 1 FPS, accommodating up to 256 frames with a maximum spatial resolution of $512 \times 512$. Correspondingly, the temporal activation window size ($W$) is set to 256.
The model is trained using the AdamW optimizer ($\beta_1=0.9$, $\beta_2=0.999$) with a global batch size of 256 and no weight decay. We apply differential learning rates: $3 \times 10^{-5}$ for the language head and $1 \times 10^{-5}$ for the language backbone, with a linear warmup of 512 steps followed by cosine decay. We use a gradient clipping threshold of 1.0, \textit{bfloat16} precision, and DeepSpeed ZeRO-2.

\begin{table}[h!]
\centering
\caption{Detailed training hyperparameters for \textit{STRIDE}.}
\label{tab:train_hparams}
\small
\resizebox{0.5\linewidth}{!}{
\begin{tabular}{@{}ll@{}}
\toprule
Config                          & Value \\
\midrule
Input frames                    & 1 FPS \\
LR scheduler                    & Linear warm-up with cosine decay \\
Warmup steps                    & 512 \\
Optimizer                       & AdamW ($\beta_1{=}0.9$, $\beta_2{=}0.999$) \\
Global Batch size               & 256 \\
Learning rate (lang. head)      & $3 \times 10^{-5}$ \\
Learning rate (lang. backbone)  & $1 \times 10^{-5}$ \\
Weight decay                    & 0 \\
Gradient clipping               & 1.0 \\
Training precision              & \textit{bfloat16} \\
DeepSpeed                       & ZeRO-2 \\
Input Resolution                & Upto $512\times512$ \\
Act. Window size ($W$)            & 256 \\
\bottomrule
\end{tabular}
}
\end{table}


For each training sample, the visual context window is constructed with a length uniformly sampled between $\max(L, 8)$ and $\min(L, 256)$ seconds, where $L$ denotes the source video length. This window is then randomly positioned along the video timeline, ensuring that the target event may or may not fall within the observable window. To allow the model to attend only to the current event of interest while disregarding previously completed events within the same window, the fixed \textit{inactive} positions from multi-event samples (\cref{sec:appendix_dataset}) are overridden onto the masked activation sequence after masking corruption. This ensures that these positions that these positions remain as \texttt{0} regardless of the applied mask. When the event of interest lies entirely outside the context window, the entire activation sequence is set to \texttt{0}, training the model not to trigger. Conversely, all activation positions that temporally overlap with the target event are set to \texttt{1}, enabling the downstream model to be invoked at the appropriate time.

\section{Detailed Architecture and Inference of STRIDE}

\subsection{Masked Diffusion Formulation for Activation Span Modeling}
\label{sec:appendix_mdm}

In \textit{STRIDE}, activation span prediction is formulated as a masked diffusion process over a discrete activation sequence defined along the video timeline. Given a video and a query describing an event of interest, the supervision signal is represented as a binary activation sequence $\textbf{\textit{a}}_0 = (a_0^1, \ldots, a_0^W)$ of length $W$, where $a_0^i \in \{0,1\}$ indicates whether the queried event is active at position $i$. Positions inside the annotated event span are labeled as $1$ (\textit{active}), while all other positions are labeled as $0$ (\textit{inactive}).

\subsubsection{Forward Corruption Process.}
To enable iterative refinement during training, we apply a masked diffusion corruption process to the activation sequence.
Starting from the ground-truth sequence $\textbf{\textit{a}}_0$, the forward process progressively masks tokens using a special symbol $\texttt{[M]}$.
At noise level $t \in [0,1]$, tokens are replaced by $\texttt{[M]}$ with probability $t$, producing a partially corrupted sequence $\textbf{\textit{a}}_t$.
The corruption process factorizes across positions:
\begin{equation}
q(\textbf{\textit{a}}_t \mid \textbf{\textit{a}}_0) = \prod_{i=1}^{W} q(a_t^i \mid a_0^i),
\quad
q(a_t^i \mid a_0^i) =
\begin{cases}
1 - t & \text{if } a_t^i = a_0^i, \\
t & \text{if } a_t^i = \texttt{[M]} .
\end{cases}
\end{equation}

Here, it is important to note that \textit{STRIDE} does not apply token-wise independent masking in practice. Instead, we use the proposed \textit{structured masking strategy} (\cref{{sec:masking}}) during training, where masking patterns are constructed to preserve contiguous temporal context along the video timeline while selectively hiding portions of the activation sequence. This encourages the model to infer coherent activation spans rather than predicting isolated token states.

\subsubsection{Reverse Denoising Process.}
The forward corruption process admits a reverse denoising process that reconstructs the clean activation sequence from a corrupted sequence. Given a partially masked activation sequence $\textbf{\textit{a}}_t$, the model predicts the original activation token at each masked position conditioned on the observed context. During the reverse process, tokens that have already been revealed remain unchanged, while masked positions are progressively unmasked according to the model’s predictions or kept masked for further refinement. Through this iterative denoising process, the model gradually recovers the activation sequence and refines activation predictions across the entire video timeline.

\subsection{Training Objective for Activation Diffusion}
The model is trained by minimizing a cross-entropy loss over masked activation positions. 
Following the masked diffusion formulation, the objective can be written as:

\begin{equation}
\mathcal{L}(\theta) =
- \mathbb{E}_{t \sim U[0,1],\, \textbf{\textit{a}}_t \sim q(\textbf{\textit{a}}_t \mid \textbf{\textit{a}}_0)}
\left[
\frac{1}{t}
\sum_{i=1}^{W}
\mathds{1}[a_t^i = \texttt{[M]}]
\log p_\theta(a_0^i \mid \textbf{\textit{a}}_t)
\right],
\label{eq:mdm_loss_appendix}
\end{equation}
where $\textbf{\textit{a}}_0$ denotes the ground-truth activation sequence and $\textbf{\textit{a}}_t$ is the corrupted sequence obtained through the forward masking process. 
The indicator function $\mathds{1}[\cdot]$ restricts the loss to masked positions only, allowing the model to leverage all unmasked tokens as context when predicting activation states. 
The $1/t$ weighting normalizes the expected number of masked positions, ensuring that the loss contribution remains balanced across different noise levels $t$.

\subsection{Inference Procedure for Activation Span Prediction}
During inference, \textit{STRIDE} predicts activation spans through an iterative reverse denoising process over the activation sequence. The process starts from a fully masked activation sequence $\textbf{\textit{a}}_1 = (\texttt{[M]}, \ldots, \texttt{[M]})$ of length $W$. Reverse denoising is then performed through $K$ discrete refinement steps following a noise schedule $1 = t_K > t_{K-1} > \cdots > t_0$. At each step from $t_k$ to $t_{k-1}$, the model predicts activation tokens for all currently masked positions conditioned on the video and query representations. A fraction $(t_k - t_{k-1})/t_k$ of masked positions with the highest confidence scores are revealed, while the remaining positions stay masked for further refinement. Positions that have already been revealed remain unchanged during the remaining denoising steps. Through this iterative predict-and-refine procedure, it progressively reconstructs the activation sequence and produces temporally consistent activation spans across the video timeline.

To support streaming inference, \textit{STRIDE} maintains the activation sequence as a sliding window over the most recent $W$ positions of the timeline. When a new frame at time $T\!+\!1$ arrives, the window shifts forward: the oldest position is removed from the window, previously resolved activations are carried forward to their shifted positions, and a new slot corresponding to the incoming frame is appended. To verify whether previously inferred activations remain valid under the updated visual context $\mathcal{V}_{\leq T+1}$, \textit{STRIDE} performs selective re-masking using a confidence threshold $\tau$. If the confidence of a carried-forward decision exceeds $\tau$, the activation is retained; otherwise, the position is re-masked as $\texttt{[M]}$ so that it re-enters the denoising process. The resulting masked set consists of both newly appended positions and low-confidence carried-forward positions, which are then refined through the same $K$-step denoising procedure described above. After the window is fully resolved, a trigger is issued only when an active span occupies at least a fraction $\gamma$ of the activation window, where $\gamma$ denotes the span ratio.

Since the activation model processes the video stream at 1 FPS, the visual context grows linearly with elapsed time. To prevent excessive computational overhead from unbounded context accumulation, we set the maximum context window size to 256 frames. When the accumulated context exceeds this limit, we retain only the most recent 128 frames and rebuild the context window from that point onward, effectively halving the temporal scope while preserving the latest visual evidence. Although cases where no trigger occurs for more than 256 consecutive seconds are rare in our evaluation benchmarks, this sliding window mechanism ensures that \textit{STRIDE} remains deployable in arbitrarily long streams without memory overflow.

\section{Detailed Benchmark and Evaluation Setup}
\subsection{Additional Benchmark Explanation}
\subsubsection{OVO-Bench.}
\label{app:ovo-bench}
OVO-Bench~\cite{niu2025ovo} evaluates temporal awareness in online video understanding by posing questions at specific timestamps during a video stream, rather than after the entire video has been observed. This timestamp-conditioned protocol reflects a core challenge of streaming scenarios: the model must reason under partial observability, where future frames are not yet available at query time. The benchmark comprises 644 videos with 2,814 human-curated question-answer pairs across 12 tasks.

The tasks are organized into three scenarios that capture distinct temporal reasoning patterns. Backward Tracing requires the model to recall and reason about past events, covering tasks such as EPM (Episodic Memory), ASI (Action Sequence Identification), and HLD (Hallucination Detection). Real-Time Visual Perception tests understanding of what is currently happening at the query timestamp, with six tasks spanning spatial understanding (STU), object recognition (OJR), attribute recognition (ATR), action recognition (ACR), optical character recognition (OCR), and future prediction (FPD). Forward Active Responding is the most distinctive scenario: the model receives a question whose answer depends on events that have not yet occurred, and must actively decide to wait rather than respond prematurely. This includes REC (Repetition Event Count), SSR (Sequential Steps Recognition), and CRR (Clues Reveal Responding).
Backward Tracing and Real-Time Visual Perception tasks adopt a multiple-choice format with accuracy as the evaluation metric. The Forward Active Responding scenario employs both accuracy-based and score-based evaluation metrics through a multiple-triggering evaluation pipeline that densely queries models along the temporal axis. This scenario is directly relevant to proactive streaming, as it requires the model to judge \textit{when} sufficient evidence has been gathered, a capability closely aligned with activation timing.

\subsubsection{StreamingBench.}
\label{app:streamingbench}
StreamingBench~\cite{lin2024streamingbench} is designed to evaluate streaming comprehension by presenting questions at different temporal positions within a video, simulating how a user might interact with a model during real-time playback. The benchmark contains 900 videos with 4,500 human-curated QA pairs (five per video), evaluated across 18 tasks under three dimensions.
Real-time Visual Understanding (10 tasks) covers a broad range of perceptual abilities including object perception (OP), causal reasoning (CR), clips summarization (CS), attribute perception (ATP), event understanding (EU), text-rich understanding (TR), prospective reasoning (PR), spatial understanding (SU), action perception (ACP), and counting (CT). These tasks collectively test whether the model can track and interpret visual changes as the stream progresses.
Omni-Source Understanding (4 tasks) requires integrating audio and visual signals, with tasks on emotion recognition (ER), scene understanding (SCU), source discrimination (SD), and multimodal alignment (MA).
Contextual Understanding (4 tasks) evaluates higher-level reasoning over accumulated context, including misleading context understanding (MCU), anomaly context understanding (ACU), sequential question answering (SQA), and proactive output (PO). The PO task is notable in that the model must determine the appropriate moment to respond without receiving an explicit user query, directly testing proactive timing capabilities. A response is considered correct only if the difference between the actual output timestamp and the ground-truth timestamp is less than two seconds.
All tasks follow a multiple-choice format with accuracy as the primary metric. Each question is evaluated on the video segment from the beginning to the timestamp when the question is asked, approximating streaming conditions.

\subsubsection{ET-Bench.}
\label{app:et-bench}
ET-Bench~\cite{liu2024bench} is a large-scale benchmark for event-level video understanding that emphasizes fine-grained temporal localization over multi-event videos. The full benchmark spans 7,300 samples across 12 tasks with 7,000 videos (251.4 hours) covering 8 domains. In our work, we evaluate on a subset of five tasks that directly measure temporal boundary prediction quality, which serves as a proxy for activation timing accuracy independent of the end-to-end streaming pipeline.

The five tasks we adopt are as follows. Temporal Video Grounding (TVG) requires localizing the temporal segment that matches a given text description within a video. Episodic Memory (EPM) extends this to egocentric scenarios, where the model must locate the moment relevant to a natural-language question (\eg, ``Where did I put my keys?''). Temporal Action Localization (TAL) involves detecting all segments containing a specified action category, testing the model's ability to identify repeated events with accurate boundaries.
Dense Video Captioning (DVC) requires jointly segmenting a video into events and generating a caption for each, evaluating both localization and description quality. Step Localization and Captioning (SLC) is similar but targets instructional videos, where the model must identify and describe sequential procedural steps.

These tasks span grounding and dense captioning capabilities under the ET-Bench taxonomy, sharing the common requirement of precise event boundary detection. To evaluate temporal localization performance, we report the F1 score as the evaluation metric for all five tasks. This metric directly measures how accurately the predicted event boundaries align with the ground-truth segments, allowing us to assess the quality of activation timing produced by the model.

\subsection{Comparison with Autoregressive Baseline}
\label{sec:appendix_baseline_ar}
To provide a fair comparison with autoregressive activation modeling, we reproduce an autoregressive baseline (Baseline-AR) following the design described in StreamBridge~\cite{wang2025streambridge}. Since the original training code and model parameters are not publicly available, we implement the baseline ourselves and train it under the same experimental setting as \textit{STRIDE}. In particular, the baseline uses the same backbone, training data, and input configuration, ensuring that the comparison primarily reflects the difference between autoregressive point-wise triggering and the proposed span-level denoising formulation.

\subsubsection{Architecture.}
We adopt Qwen3-VL 2B as the backbone, processing up to 256 frames at 1 FPS to match the input configuration used in \textit{STRIDE}. Following StreamBridge~\cite{wang2025streambridge}, a learnable \texttt{<ACT>} token is appended after the visual embedding of each frame. The token is passed through a lightweight score head that performs binary classification to predict whether a trigger should be issued at the corresponding time step.

\subsubsection{Training \& Inference.}
The training data is constructed from the same annotations used for training \textit{STRIDE}. For each annotated event segment, the last $P$\% of frames within the segment are labeled as \textit{active} (trigger-positive), while all remaining frames are labeled as \textit{inactive}. To expose the model to diverse activation patterns, $P$ is randomly sampled from a uniform distribution over $[0, 50]$ for each training sample, following the training protocol of StreamBridge~\cite{wang2025streambridge}.

At inference time, the score head outputs a trigger probability for each frame independently. A fixed threshold of 0.35 is applied to determine whether a trigger should be issued at each time step, following StreamBridge~\cite{wang2025streambridge}. This point-wise decision mechanism contrasts with \textit{STRIDE}, which jointly denoises the activation sequence to produce span-level activation predictions.

\section{Additional Experimental Analysis}
\subsection{Sensitivity Analysis for $\tau$}
In a streaming setting, the activation window slides as new visual context arrives. To re-evaluate previously resolved decisions against new visual evidence, we apply a confidence-based retention threshold $\tau$ (see Selective Re-masking in \cref{fig:2}). Here, $\tau\!=\!0$ unconditionally retains all prior decisions, while $\tau\!=\!1$ effectively rebuilds the activation window from scratch at every step. To validate the effect of the Selective Re-masking threshold $\tau$ (\cref{sec:method}), we evaluate performance on ET-Bench by sweeping $\tau$ from 0 to 1.

\begin{figure*}[h!]
\centering
\includegraphics[width=0.8\linewidth]{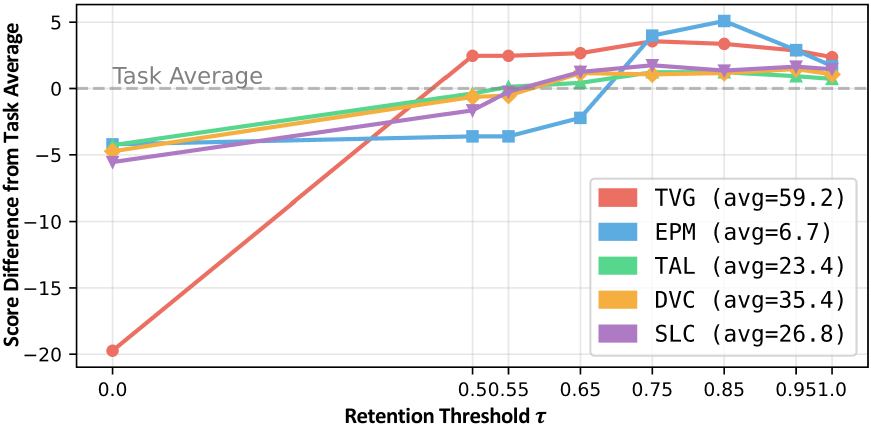}
\caption{Sensitivity of \textit{STRIDE} to the \textit{retention constant} $\tau$ across five temporal understanding tasks (TVG, EPM, TAL, DVC, SLC) in ET-Bench~\cite{liu2024bench}. The $y$-axis shows the score difference relative to each task's average. Task-wise average scores are shown in the legend.}
\label{fig:confidence}
\end{figure*}

As shown in \cref{fig:confidence}, unconditional inheritance ($\tau\!=\!0$) results in the lowest scores, with a particularly large drop of $-19.7$ pt on TVG. Performance peaks broadly in the range $\tau \in [0.75, 0.85]$ across most tasks, after which tightening the retention criterion causes a gradual decline. Based on these observations, $\tau\!=\!0.75$ is used in all evaluations.

\subsection{Scalability Analysis for Activation Backbone}
To examine how \textit{STRIDE} scales with different size activation backbones, we conduct an additional experiment by replacing the default Qwen3-VL 2B activation backbone with a larger size Qwen3-VL 4B model. The 4B variant is trained nder the same data and training configuration as the 2B model and evaluated across three downstream Video-LLMs~\cite{team2025gemma,wang2025internvl3,Qwen3-VL} on OVO-Bench (\cref{tab:scale_ovo}) and StreamingBench (\cref{tab:scale_streaming}).

Across all downstream backbones, \textit{STRIDE}-4B consistently achieves higher overall scores than \textit{STRIDE}-2B, confirming that the activation backbone benefits from increased model capacity and that the improvement transfers regardless of the downstream Video-LLM, supporting the scalability of the proposed plug-in design.
\begin{table}[h!]
\centering
\small
\setlength{\tabcolsep}{3.5pt}
\caption{Scalability analysis on activation backbone scale (\textit{STRIDE}-2B vs. 4B) on OVO-Bench~\cite{niu2025ovo} across multiple downstream Video-LLMs.}
\label{tab:scale_ovo}
\resizebox{1.0\linewidth}{!}{
\begin{tabular}{@{}l cccccc c !{\vrule} ccc c !{\vrule} ccc c !{\vrule} c@{}}
\Xhline{2\arrayrulewidth}
\multirow{2}{*}[-4pt]{Method} & \multicolumn{7}{c}{\textbf{Real-Time Visual Perception}} & \multicolumn{4}{c}{\textbf{Backward Tracing}} & \multicolumn{4}{c}{\textbf{Fwd. Act. Responding}} & \textbf{Overall} \\
\cmidrule(l{2pt}r{2pt}){2-8} \cmidrule(l{2pt}r{2pt}){9-12} \cmidrule(l{2pt}r{2pt}){13-16} \cmidrule(l{2pt}r{2pt}){17-17}
& OCR & ACR & ATR & STU & FPD & OJR & Avg. & EPM & ASI & HLD & Avg. & REC & SSR & CRR & Avg. & Avg. \\
\hline
Gemma3-4B~\cite{team2025gemma}              & 65.8 & 48.6 & 56.0 & 36.0 & 66.3 & 50.0 & 53.78 & 44.4 & 41.9 & 3.2  & 29.83 & 14.4 & 61.4 & 52.5 & 42.77 & 42.13 \\
\hspace{0.5em}+ \textbf{\textit{STRIDE}}-2B         & 73.2 & 60.6 & 64.7 & 39.3 & 71.3 & 56.5 & 60.93 & 47.8 & 52.0 & 4.8  & 34.87 & 42.6 & 64.6 & 60.0 & 55.73 & 50.51 \\
\hspace{0.5em}+ \textbf{\textit{STRIDE}}-4B         & 75.2 & 56.9 & 67.2 & 40.4 & 67.3 & 56.5 & 60.58 & 51.5 & 48.0 & 4.3  & 34.60 & 46.5 & 66.2 & 60.0 & 57.57 & \textbf{50.92} \\
\hdashline
InternVL3-8B~\cite{wang2025internvl3}       & 65.8 & 52.3 & 68.1 & 51.1 & 71.3 & 62.0 & 61.77 & 58.9 & 66.9 & 9.7  & 45.17 & 36.6 & 64.1 & 43.3 & 48.00 & 51.64 \\
\hspace{0.5em}+ \textbf{\textit{STRIDE}}-2B         & 75.8 & 54.1 & 80.2 & 56.7 & 74.3 & 65.2 & 67.72 & 58.9 & 65.5 & 11.3 & 45.23 & 40.1 & 67.7 & 66.2 & 58.00 & 56.98 \\
\hspace{0.5em}+ \textbf{\textit{STRIDE}}-4B & 78.5 & 56.9 & 75.0 & 54.5 & 71.3 & 63.6 & 66.63 & 59.6 & 67.6 & 16.1 & 47.77 & 44.4 & 68.4 & 59.2 & 57.33 & \textbf{57.24} \\
\hdashline
Qwen3-VL-8B~\cite{Qwen3-VL}                & 69.8 & 59.6 & 73.3 & 57.3 & 71.3 & 58.7 & 65.00 & 55.6 & 63.5 & 12.9 & 44.00 & 37.7 & 60.8 & 40.4 & 46.30 & 51.77 \\
\hspace{0.5em}+ \textbf{\textit{STRIDE}}-2B         & 76.5 & 64.2 & 79.3 & 61.2 & 73.3 & 63.6 & 69.68 & 57.2 & 72.3 & 14.0 & 47.83 & 46.4 & 63.1 & 69.6 & 59.70 & 59.07 \\
\hspace{0.5em}+ \textbf{\textit{STRIDE}}-4B & 75.2 & 64.2 & 76.7 & 62.9 & 73.3 & 66.3 & 69.77 & 59.3 & 74.3 & 12.4 & 48.67 & 57.1 & 64.6 & 65.4 & 62.37 & \textbf{60.27} \\
\Xhline{2\arrayrulewidth}
\end{tabular}
}
\end{table}

\begin{table*}[h!]
\centering
\small
\setlength{\tabcolsep}{3.5pt}
\caption{Scalability analysis on activation backbone scale (\textit{STRIDE}-2B vs. 4B) on StreamingBench~\cite{lin2024streamingbench} across multiple downstream Video-LLMs.}
\label{tab:scale_streaming}
\resizebox{1.0\linewidth}{!}{
\begin{tabular}{@{}l cccccccccc c !{\vrule} cccc c !{\vrule} cccc c !{\vrule} c@{}}
\Xhline{2\arrayrulewidth}
\multirow{2}{*}[-4pt]{Method} & \multicolumn{11}{c}{\textbf{Real-Time Visual Understanding}} & \multicolumn{5}{c}{\textbf{Omni-Source Understanding}} & \multicolumn{5}{c}{\textbf{Contextual Understanding}} & \textbf{Overall} \\
\cmidrule(l{2pt}r{2pt}){2-12} \cmidrule(l{2pt}r{2pt}){13-17} \cmidrule(l{2pt}r{2pt}){18-22} \cmidrule(l{2pt}r{2pt}){23-23}
& OP & CR & CS & ATP & EU & TR & PR & SU & ACP & CT & Avg. & ER & SCU & SD & MA & Avg. & ACU & MCU & SQA & PO & Avg. & Avg. \\
\hline
Gemma3-4B~\cite{team2025gemma}              & 63.8 & 69.5 & 68.8 & 54.4 & 60.2 & 65.1 & 59.3 & 40.7 & 62.7 & 21.8 & 57.49 & 28.8 & 31.2 & 30.4 & 46.4 & 34.20 & 34.4 & 31.2 & 38.8 & 12.4 & 29.20 & 40.30 \\
\hspace{0.5em}+ \textbf{\textit{STRIDE}}-2B         & 66.8 & 71.9 & 66.6 & 57.2 & 66.5 & 70.7 & 60.2 & 43.1 & 65.0 & 23.8 & 60.00 & 35.6 & 31.6 & 36.0 & 44.0 & 36.80 & 33.6 & 35.2 & 44.8 & 41.6 & 38.80 & 45.20 \\
\hspace{0.5em}+ \textbf{\textit{STRIDE}}-4B         & 66.2 & 63.3 & 68.1 & 58.1 & 67.1 & 71.7 & 63.0 & 41.5 & 66.7 & 21.2 & 59.93 & 32.8 & 30.4 & 36.0 & 46.4 & 36.40 & 35.2 & 36.8 & 48.0 & 44.0 & 41.00 & \textbf{45.78} \\
\hdashline
InternVL3-8B~\cite{wang2025internvl3}       & 74.9 & 82.0 & 75.7 & 61.2 & 72.0 & 67.6 & 74.1 & 66.7 & 78.1 & 34.2 & 68.71 & 40.4 & 27.6 & 38.8 & 45.6 & 38.10 & 38.0 & 26.0 & 36.8 & 31.2 & 33.00 & 46.60 \\
\hspace{0.5em}+ \textbf{\textit{STRIDE}}-2B         & 74.9 & 78.9 & 76.7 & 68.6 & 77.0 & 77.3 & 77.8 & 71.5 & 83.0 & 33.2 & 72.45 & 39.6 & 22.4 & 44.0 & 50.8 & 39.20 & 34.0 & 35.2 & 43.2 & 42.8 & 38.80 & 50.15 \\
\hspace{0.5em}+ \textbf{\textit{STRIDE}}-4B         & 77.7 & 81.2 & 79.5 & 70.0 & 75.2 & 77.3 & 73.1 & 72.4 & 84.3 & 37.8 & 73.82 & 38.8 & 25.6 & 43.6 & 55.6 & 40.90 & 39.2 & 35.2 & 44.4 & 44.8 & 40.90 & \textbf{51.87} \\
\hdashline
Qwen3-VL-8B~\cite{Qwen3-VL}                & 62.7 & 68.0 & 69.7 & 53.3 & 67.5 & 65.1 & 67.6 & 48.0 & 68.0 & 40.9 & 60.88 & 36.8 & 18.0 & 32.8 & 34.0 & 30.40 & 25.6 & 23.6 & 31.2 & 32.4 & 28.20 & 39.83 \\
\hspace{0.5em}+ \textbf{\textit{STRIDE}}-2B         & 77.1 & 75.0 & 77.3 & 72.8 & 76.4 & 77.9 & 76.9 & 69.9 & 84.3 & 46.1 & 74.24 & 42.8 & 24.0 & 45.2 & 53.2 & 41.30 & 32.0 & 38.4 & 46.4 & 42.8 & 39.90 & 51.81 \\
\hspace{0.5em}+ \textbf{\textit{STRIDE}}-4B         & 80.7 & 78.9 & 79.8 & 73.1 & 78.9 & 84.1 & 77.8 & 69.9 & 84.0 & 42.5 & 76.01 & 42.8 & 21.2 & 41.6 & 54.4 & 40.00 & 36.0 & 36.8 & 40.8 & 46.0 & 39.90 & \textbf{51.97} \\
\Xhline{2\arrayrulewidth}
\end{tabular}
}
\end{table*}


\subsection{Qualitative Examples of STRIDE}
We provide qualitative examples of activation span prediction on OVO-Bench, StreamingBench, and ET-Bench in \cref{fig:qual-assets/tvg_ZjFzkhrqIZs,fig:qual-assets/tvg_czN-9IsQXoU,fig:qual-assets/epm_8a89601b,fig:qual-assets/tal_video_test_0000737,fig:qual-assets/dvc_LfSYF1N5i_Q,fig:qual-assets/slc_ejdn67zHuEY,fig:qual-assets/ovo_EPM_4a37144f,fig:qual-assets/ovo_OJR_247,fig:qual-assets/ovo_OJR_181,fig:qual-assets/sb2_sample_308_Attribute_Recognition_05,fig:qual-assets/sb2_sample_308_Object_Recognition_03,fig:qual-assets/sb2_sample_119_Text-Rich_Understanding_04}. Each example visualizes the temporal timeline of the video together with the query arrival time, the ground-truth event span, and the activation span predicted by \textit{STRIDE}. 
These timelines illustrate how the model progressively identifies the relevant event segment and aligns its activation predictions with the ground-truth temporal boundaries under streaming conditions.

\section{Limitation and Discussion}
\subsection{Failure Cases and Discussion}
Although \textit{STRIDE} improves temporal stability of activation decisions, several practical limitations remain in streaming deployments. First, the activation model operates on sparsely sampled frames (1 FPS) and relies on downstream Video-LLMs whose streaming interfaces typically process visual tokens at relatively low frame rates. As a result, extremely short-lived events or rapid visual transitions may not be fully captured by the activation window, since the visual evidence may disappear before sufficient temporal context is accumulated. \cref{fig:qual-assets/sb2_sample_49_Proactive_Output_04} illustrates such a case, where a brief visual event occurs between sampled frames and therefore cannot be reliably localized by the activation model.

Another challenging scenario arises when queries refer to broad or loosely defined events rather than a single well-localized moment. In such cases, multiple candidate segments may partially satisfy the query semantics, leading to dispersed or multi-span activations. \cref{fig:qual-assets/sb2_sample_49_Scene_Understanding_05} presents an example where the model encounters several visually plausible moments corresponding to the query, which may introduce ambiguity in determining the most appropriate triggering point. These observations suggest that proactive activation remains sensitive to both temporal sampling granularity and query specificity, highlighting directions for future improvements in streaming perception and query grounding.

\RenderETExampleFigure{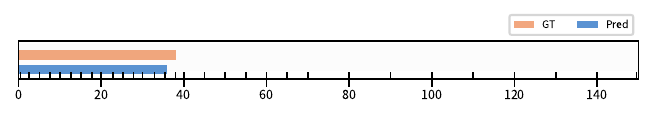}{Qualitative example from ET-Bench (TVG).}
\RenderETExampleFigure{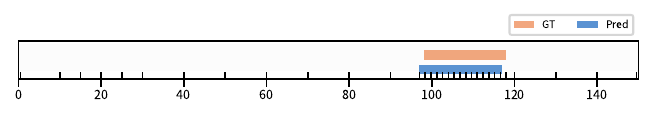}{Qualitative example from ET-Bench (TVG).}
\RenderETExampleFigure{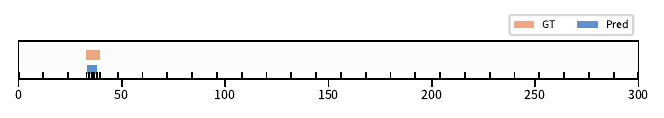}{Qualitative example from ET-Bench (EPM).}
\RenderETExampleFigure{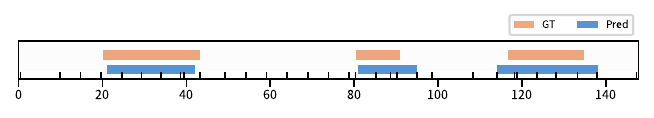}{Qualitative example from ET-Bench (TAL).}
\RenderETExampleFigure{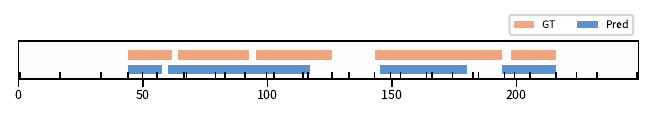}{Qualitative example from ET-Bench (DVC).}
\RenderETExampleFigure{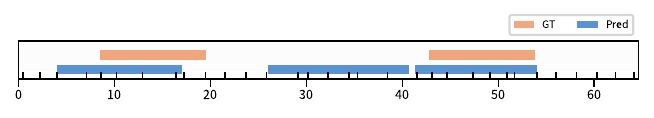}{Qualitative example from ET-Bench (SLC).}

\RenderQAExampleFigure[0.6]{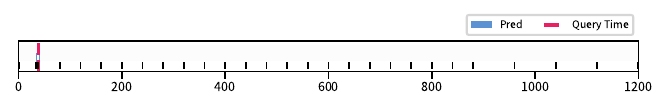}{Qualitative example from OVO-Bench (EPM).}
\RenderQAExampleFigure{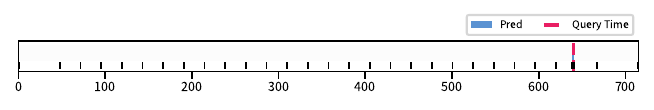}{Qualitative example from OVO-Bench (OJR).}
\RenderQAExampleFigure{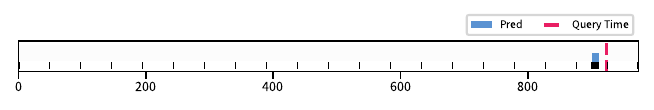}{Qualitative example from OVO-Bench (OJR).}

\RenderQAExampleFigure{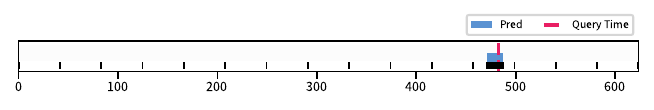}{Qualitative example from StreamingBench (Attribute Recognition).}
\RenderQAExampleFigure{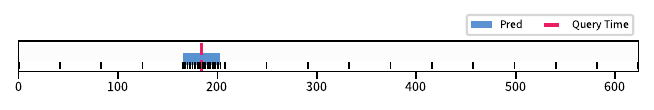}{Qualitative example from StreamingBench (Object Recognition).}
\RenderQAExampleFigure{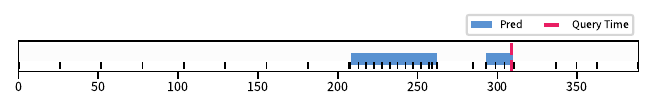}{Qualitative example from StreamingBench (Text-Rich Understanding).}
\RenderETExampleFigure{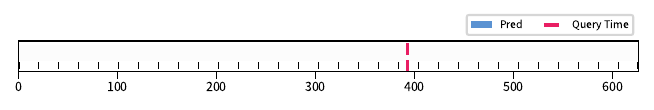}{Failure Case from StreamingBench (Proactive Output).}
\RenderQAExampleFigure{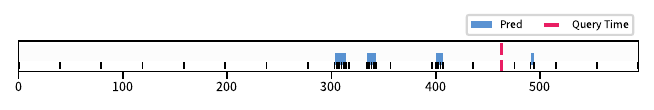}{Failure Case from StreamingBench (Scene Understanding).}

\end{document}